\documentclass[10pt,twocolumn,letterpaper]{article}

\usepackage{cvpr}
\usepackage{times}
\usepackage{epsfig}
\usepackage{graphicx}
\usepackage{amsmath}
\usepackage{amssymb}
\usepackage{dsfont}
\usepackage{cite}
\usepackage{subfigure}
\usepackage{wrapfigure}

\usepackage[bookmarks=false]{hyperref}

\cvprfinalcopy 

\def\cvprPaperID{****} 

\begin{document}

\title{3D Semantic Trajectory Reconstruction from 3D Pixel Continuum}

\author{Jae Shin Yoon\\
University of Minnesota\\
{\tt\small yoon0074@umn.edu}
\and
Ziwei Li\\
University of Minnesota\\
{\tt\small lixx3686@umn.edu}
\and
Hyun Soo Park\\
University of Minnesota\\
{\tt\small hspark@umn.edu}
}
\maketitle

\begin{abstract}
	This paper presents a method to reconstruct dense semantic trajectory stream of human interactions in 3D from synchronized multiple videos. The interactions inherently introduce self-occlusion and illumination/appearance/shape changes, resulting in highly fragmented trajectory reconstruction with noisy and coarse semantic labels. Our conjecture is that among many views, there exists a set of views that can confidently recognize the visual semantic label of a 3D trajectory. We introduce a new representation called 3D semantic map---a probability distribution over the semantic labels per trajectory. We construct the 3D semantic map by reasoning about visibility and 2D recognition confidence based on view-pooling, i.e., finding the view that best represents the semantics of the trajectory. Using the 3D semantic map, we precisely infer all trajectory labels jointly by considering the affinity between long range trajectories via estimating their local rigid transformations. This inference quantitatively outperforms the baseline approaches in terms of predictive validity, representation robustness, and affinity effectiveness. We demonstrate that our algorithm can robustly compute the semantic labels of a large scale trajectory set involving real-world human interactions with object, scenes, and people. 
	
\end{abstract}

\section{Introduction}
\begin{figure}[t]
	\centering  
	\includegraphics[width=0.45\textwidth]{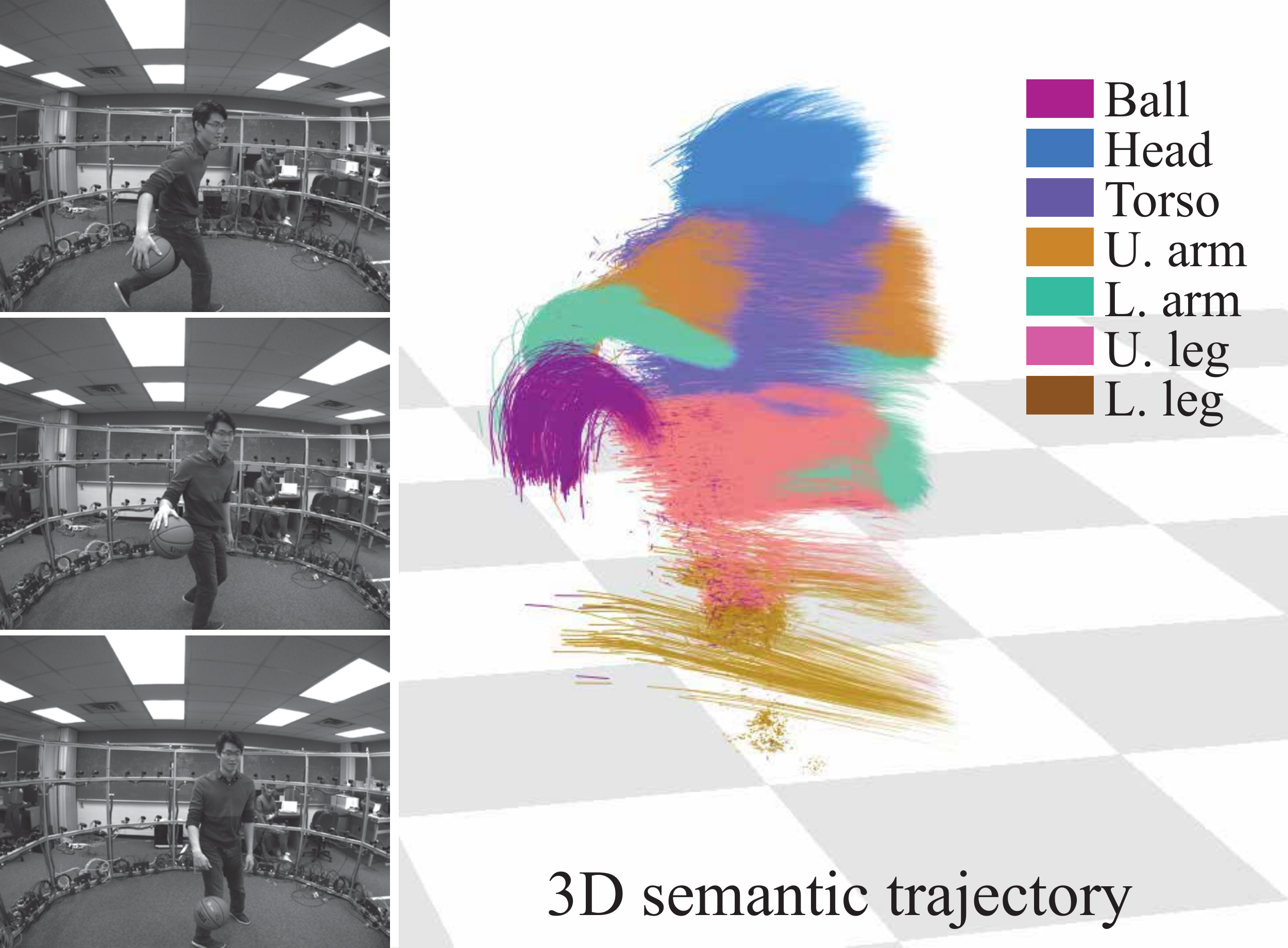}
	\caption{We reconstruct 3D dense semantic trajectories in 3D using a large scale multicamera system. Each trajectory is associated with semantic labels such as body parts and objects (basketball). For illustrative purpose, the last 10 frames of trajectories are visualized.} 
	\label{Fig:teaser}
\end{figure}

Now cameras are deeply integrated in our daily lives, e.g., Amazon Cloud Cam and Nest Cam, reaching soon towards {\em 3D pixel continuum}---every 3D point in our space is observed in a form of multiple view pixels by a network of ubiquitous cameras. Such cameras open up a unique opportunity to quantitatively analyze our detailed interactions with scenes, objects, and people continuously, which will facilitate behavioral monitoring for the elderly, human-robot collaboration, and social tele-presence. A 3D trajectory representation of human interactions~\cite{bregler:2000,torresani:2001,ozden:2004,yan:2006,joo_cvpr_2014} is a viable computational model that measures microscopic actions at high spatial resolution without prior scene assumptions. Unfortunately, the representation is a lack of semantics, which fundamentally prevents from computational behavioral analysis. It is important to know not only where a 3D point is but also what it means and how associated with other points. For instance,  as shown in Figure~\ref{Fig:teaser}, the trajectory of the basketball player's hand (semantics) is spatially and temporally related with another trajectory of the ball to encode their physical interactions.  


However, assigning a semantic label to each trajectory in the wild involves with two principal challenges. (1) Missing data: interactions with objects and others inherently introduce significant occlusion, resulting in fragmented trajectories, i.e., each trajectory emerges and dissolves in different time instances where prior approaches of global spatial reasoning such as articulated body~\cite{yan:2006} and shape basis~\cite{bregler:2000,torresani:2001} are not applicable. Occlusion further introduces the label ambiguity where multiple labels may be associated with a single 3D trajectory. (2) Noisy and coarse recognition: existing visual recognition systems were largely built on single perspective images, which are often fragile to heavy background clutter, self-occlusion, and non-canonical object pose. This problem escalates when low resolution models such as a bounding box representation are used where not all pixels in a detection window belong to the same object class.

In this paper, we present a method to precisely reconstruct dense semantic trajectories in 3D by leveraging a multicamera system that emulates the 3D pixel continuum. Our method uses two cues. (a) 2D visual cue: albeit noisy, a series of recognition results across multiple views can be consolidated, e.g., among many views, there exists a set of views that can confidently recognize the label of a 3D trajectory. We introduce a new representation called {\em 3D semantic map}---a probability distribution over semantic labels per 3D trajectory. We construct the 3D semantic map based on visibility and recognition confidence. (b) 3D spatial cue: if trajectories are sufficiently dense, a set of trajectories that belong to the same objects can be expressed by local rigid transformation. This allows computing an affinity measure between long range fragmented trajectories. To achieve that, we reconstruct 3D trajectories in unprecedented resolution (e.g., $>100,000$ per object for each second) in aid of dense image matching. 


Our system takes a set of synchronized image streams captured by 69 HD cameras\footnote{Our system reaches average 6.4 pixels/cm$^3$, resulting in the most dense 3D pixel continuum. cf) 0.44 pixels/cm$^3$ for the Panoptic Studio at CMU~\cite{joo_cvpr_2014,joo_iccv_2015}}. At each time instant, dense 3D points are reconstructed and tracked across time, which forms a set of long term trajectories. We build the 3D semantic map using view-pooling that reasons about visibility and recognition confidence. This allows to find the view that best represents the semantics of a 3D trajectory. Using the 3D semantic map, we precisely infer all trajectory labels jointly by considering the affinity between long range trajectories
via estimating local rigid transformations. The inference is conducted via multi-class graph-cuts in Markov Random Field (MRF).


The core contributions of this paper include: (1) 3D semantic map: we introduce a novel concept for trajectory semantics encoding the distribution over labels, which can be computed by view-pooling; (2) Long range affinity: estimation of local rigid transformation around a trajectory allows relating with distant trajectories; (3) Multiple view human interaction dataset: we collect 9 new datasets involving in various human interactions including pet/social interactions, dance, sports, and object manipulations; (4) Modular design of 3D pixel continuum: we design a space that can densely measure human interactions from nearly exhaustive views by modularizing commodity parts, which is scalable and customizable.   

\begin{figure*}[t]
	\centering  
	\subfigure[Geometry]{\label{Fig:geom}\includegraphics[height=0.14\textheight]{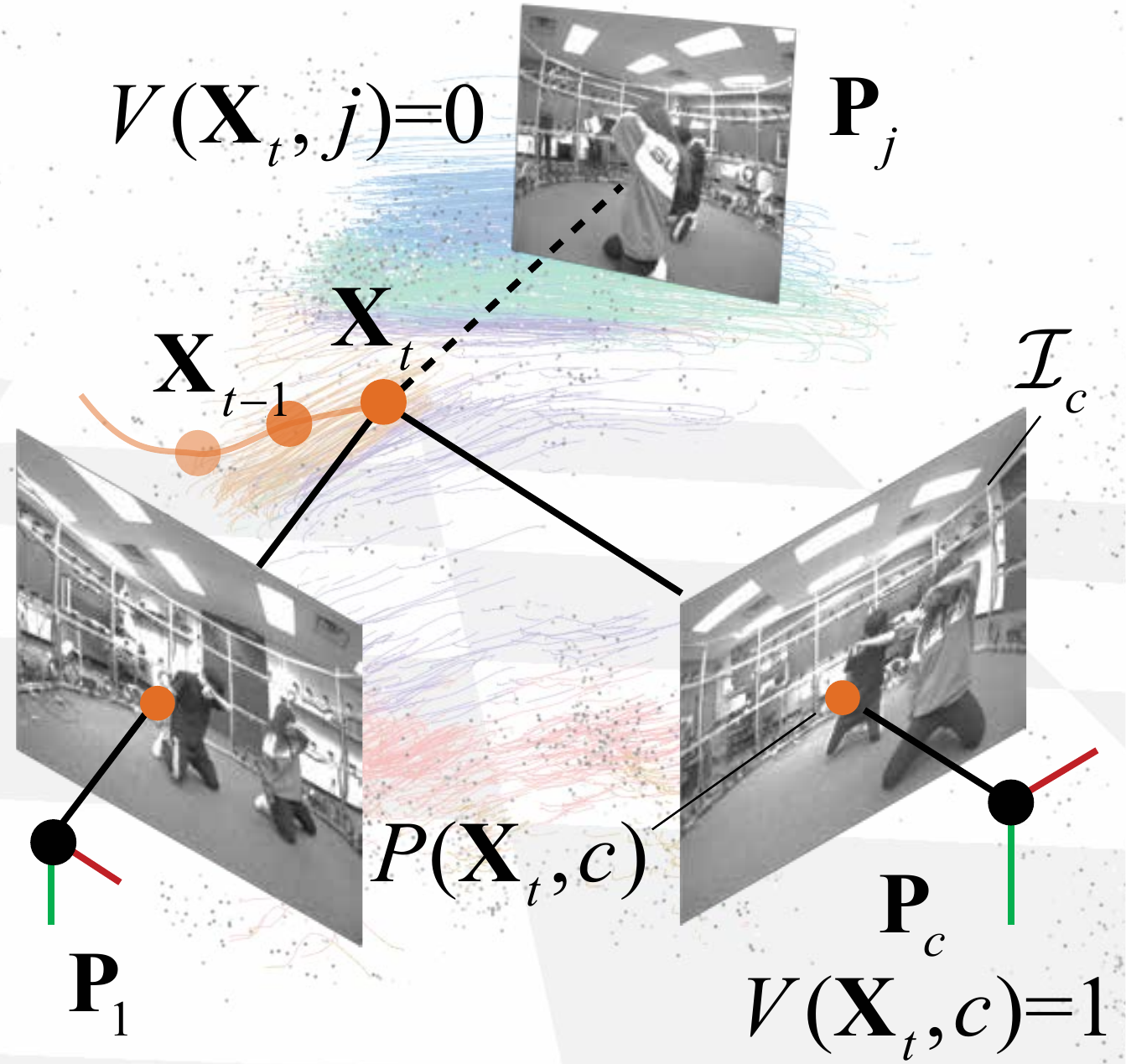}}~
	\subfigure[2D multiple view recognition]{\label{Fig:detection}\includegraphics[height=0.14\textheight]{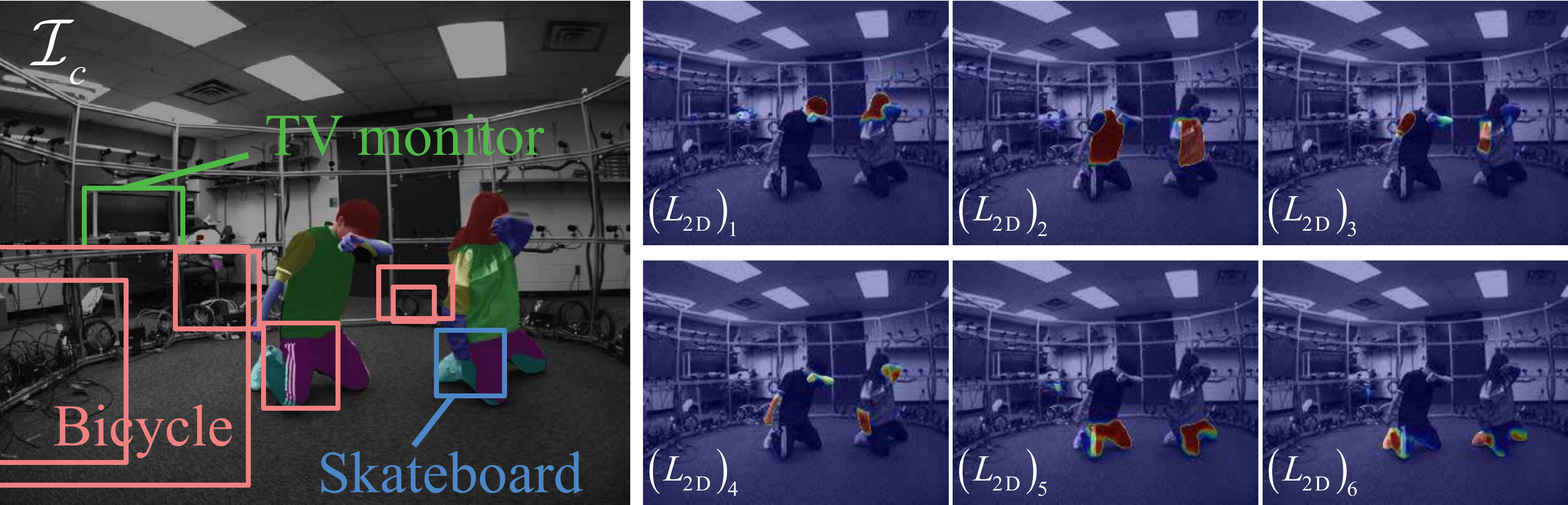}}~
	\subfigure[View-pooling]{\label{Fig:pooling}\includegraphics[height=0.14\textheight]{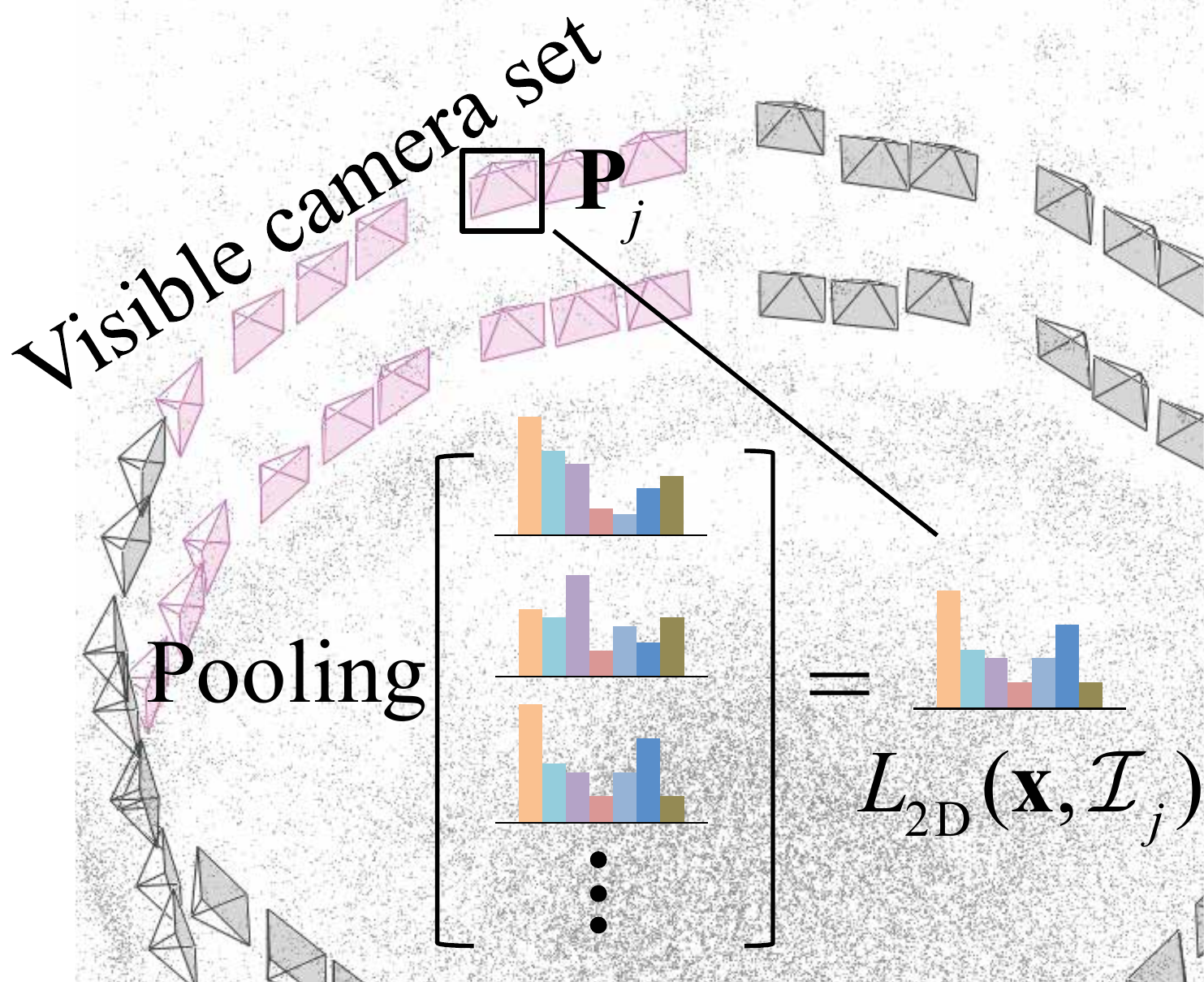}}
	\caption{(a) A 3D point $\mathbf{X}_t$ at the $t$ time instant is observed by multiple cameras $\{\mathbf{P}_c\}_{c\in\mathcal{C}}$ where the point is fully visible to the $c^{\rm th}$ camera if $V(\mathbf{X}_t,c)=1$, and zero otherwise. We denote the 2D projection of the 3D point onto the camera as $P(\mathbf{X}_t,c)$. (b) For each image $\mathcal{I}_c$, we use the recognition confidence (body segmentation~\cite{lin:2017}/object bounding box~\cite{redmon:2017}) to build $L_{2D}(\mathbf{x}|\mathcal{I}_c)$ at each pixel $\mathbf{x}$ where the $i^{\rm th}$ element of $L_{2D}$ is the likelihood (confidence) of the recognition for the $i^{\rm th}$ object class as shown on the right. For the illustration purpose, we only visualize the likelihood of body segments overlaid with the image while $L_{2D}$ also includes object classes. (c) We construct the 3D semantic map $L_{3D}(\mathcal{X})$ via pooling $L_{2D}$ over multiple views (view-pooling) by reasoning about visibility. The magenta camera is the visible camera set, and the bar graphs represent $L_{2D}$. The figures are best seen in color.} 
	\label{Fig:semantic_map}
\end{figure*}
\section{Related Work}

Humans can effortlessly {\em read} the intent of others through subtle behavioral cues in a fraction of second~\cite{ambady:1992}, and high resolution videos are now able to capture such cues via our interactions with surrounding environments. The pixels in the videos can be tracked to form long term trajectories to encode the interactions both in 2D and 3D.

\noindent\textbf{2D trajectory} As many objects are roughly rigid and move independently, motion provides a strong discriminative cue  to group pixels and recognize occluding boundary, precisely. A core challenge of motion segmentation lies in fragmented nature of trajectories caused by tracking failure (occlusion, drifting, and motion blur). Embedding trajectories into low dimensional space has been used to robustly measure trajectory distance in the presence of missing data without pre-trained models~\cite{brox:2010,fragkiadaki:2012,ricco:2013,elhamifar:2009}, and 2D trajectories can be decomposed into 3D camera motion and deformable object models~\cite{torresani:2002,rao:2010,sheikh:2009}. Visual semantics learned by object recognition frameworks provides stronger cues to cluster trajectories~\cite{taylor:2013,kundu:2014,kundu:2016}.















\noindent\textbf{3D trajectories} Due to dimensional loss in the process of 2D projection, reconstructing 3D motion from a monocular camera is an ill-posed problem in general, i.e., the number of variables (3D motion parameters) is greater than the number equations (projections). However, when an object undergoes constrained deformation such as face, its 3D shape can be recovered by enforcing spatial regularity, e.g., shape basis ~\cite{bregler:2000,torresani:2001,shaji:2010,yan:2006}, template~\cite{salzmann:2007}, and mesh~\cite{taylor:2010}. A key challenge of this approach is to learn a shape prior that can express general deformation, often requiring an instance specific pre-trained model, or inherent rank minimization where the global solution is difficult to be achieved~\cite{akhter:2009,dai:2012}. A trajectory based representation directly addresses this challenge. Motion is described by a set of trajectory stream where generic temporal regularity is applied through DCT trajectory basis~\cite{akhter:2008,park:2010}, polynomial basis~\cite{avidan:2000,kaminski:2004}, and linear dynamical model~\cite{sidenbladh:2000}. A spatiotemporal constraint can further reduce dimensionality, resulting in robust 3D reconstruction~\cite{torresani:2002,park:2011,Akhter:2012:BilinearBasis}. When multiple view images are used, it is possible to represent general motion with topological change without any spatial and temporal prior~\cite{joo_cvpr_2014,joo_iccv_2015}.  

Unlike 2D trajectories, semantic labeling of 3D trajectories is largely under-studied research area. Notably, Yan and Pollefeys~\cite{yan:2006} presented a trajectory clustering algorithm based on articulated body structure, i.e., an object is composed of a kinematic chain of rigid bodies where the articulated joint and its rotational axis lie in the intersection of two shape subspaces. Later, image segmentation cues have been incorporated to recognize a scene topology, i.e., pre-clustering object instances, to reconstruct dynamics scenes from videos in the wild~\cite{bue:2007,russell:2014,nrsfmFragkiadaki}. Note that none of these work has addressed semantics. The work by Joo et al.~\cite{joo_iccv_2015} is closest to our approach where the trajectory clustering is based on 3D rigid transformation of human anatomical keypoints. Our method is not limited to human bodies, which enables modeling general human interactions with scenes, objects, and other people.


\section{System Overview}
Our system takes 69 synchronized image streams at 30 Hz from a multicamera system (Section~\ref{Sec:continuum}). We use the standard structure from motion pipeline~\cite{hartley:2004,SNAVELY-IJCV08} to calibrate the camera and reconstruct  trajectory stream in 3D as described in Section~\ref{Sec:reconstruction}. The 3D reconstructed trajectories are used to infer their semantic labels by consolidating 2D recognition confidence in multiple view images: 3D semantic map is constructed using view-pooling (Section~\ref{Sec:map}), and affinity between long range fragmented trajectories is measured by computing local transformation (Section~\ref{Sec:aff}). The system outputs the 3D dense semantic trajectories that consistently aligns with image visual semantic recognition.  

\section{Notation}


We represent a fragmented trajectory with a time series of 3D points: $\mathcal{X} = \{\mathbf{X}_t\in \mathds{R}^3\}_{t=T_e}^{T_d}$ where $\mathbf{X}_t$ is the 3D point in the trajectory at the $t$ time instant, and $T_e$ and $T_d$ are emerging and dissolving moments of the trajectory, respectively. We denote the probability of visibility as $V(\mathbf{X}_t, c) \in [0,1]$ as shown in Figure~\ref{Fig:geom} where $c\in \mathcal{C}$ is the camera index, and $\mathcal{C}$ is the camera index set, i.e., $|\mathcal{C}|$ is the number of cameras.



The 3D point $\mathbf{X}_t$ is projected onto the visible $c^{\rm th}$ camera projection matrix, $\mathbf{P}_c = \mathbf{K}_c \mathbf{R}_c \left[\begin{array}{cc}\mathbf{I}_3 & -\mathbf{C}_c\end{array}\right] \in \mathds{R}^{3\times4}$ to form the 2D projection, $P(\mathbf{X}_t,c) \in \mathds{R}^2$ where $\mathbf{K}_c$ is the intrinsic parameter of the camera encoding focal length and principal points, and $\mathbf{R}_c \in SO(3)$ and $\mathbf{C}_c \in \mathds{R}^3$ are the extrinsic parameters (rotation and camera center), i.e., 
$P(\mathbf{X}_t,c) = \left[\begin{array}{cc}\mathbf{P}^1_c \widetilde{\mathbf{X}}_t/\mathbf{P}^3_c \widetilde{\mathbf{X}}_t& \mathbf{P}^2_c \widetilde{\mathbf{X}}_t/\mathbf{P}^3_c \widetilde{\mathbf{X}}_t\end{array}\right]^\mathsf{T}$ where $\widetilde{\mathbf{X}}$ is the homogeneous representation of $\mathbf{X}$, and $\mathbf{P}^i_c$ indicates the $i^{\rm th}$ row of $\mathbf{P}_c$. We assume the camera extrinsic and intrinsic parameters are pre-calibrated and constant across time (no time index). 

The $c^{\rm th}$ camera produces the image at the $t$ time instant $\mathcal{I}_t^c$. Each pixel $\mathbf{x}$ is associated with the confidence of semantic labels, i.e., $L_{2D}\left(\mathbf{x} \in \mathds{R}^2|\mathcal{I}_c\right) \in [0,1]^N$ where $N$ is the number of object classes\footnote{The object classes include objects, body parts, and independent instances.}. For instance, $L_{2D}$ can be approximated by the last layers of a convolutional neural network as shown in Figure~\ref{Fig:detection}. Our framework can build on general 2D recognition framework that can produce a confidence map while in this paper, we focus on two main pre-trained models: body semantic segmentation~\cite{lin:2017} and bounding box object recognition~\cite{redmon:2017}.



\section{Semantic Trajectory Labeling}
Given 3D reconstructed trajectories, we present a method to precisely infer their semantic labels. A key innovation is the {\em 3D semantic map} that can encode the visual semantics of a 3D trajectory by consolidating the 2D recognition confidence across multiple view image streams. We integrate the 3D semantic map in conjunction with long term affinity into a graph-cut formulation to infer the semantic labels jointly.



\subsection{3D Semantic Map} \label{Sec:map}
We define the 3D semantic map, $L_{3D} \in [0,1]^N$, a probability distribution over semantic labels of a 3D trajectory. It is computed by reasoning about visibility and 2D recognition confidence at the 2D projections of the trajectory onto all cameras:
\begin{align}
	&L_{3D} \left(\mathcal{X}\right) 
	= \frac{1}{\Delta T} \sum_{t=T_e}^{T_d} \underset{c \in \mathcal{C}}{\operatorname{Pool}} \left( L_{2D} \left(P\left(\mathbf{X}_t,c\right)|\mathcal{I}_c\right)\right),
\end{align}
where $\Delta T = T_d-T_e$ is the life span of the trajectory. The 3D trajectory label is evaluated at the 2D projection $P\left(\mathbf{X}_t,c\right)$ across all cameras over the trajectory life span. To alleviate noisy and coarse  2D recognition results, we introduce a view-pooling operation:
\begin{align}
	L_{c^\ast} = \underset{c\in \mathcal{C}}{\operatorname{Pool}} \left(L_c\right) \nonumber~~~~{\rm s.t.}~~
	c^\ast  =  \underset{c \in \mathcal{C}}{\operatorname{argmin}}~~ \sum_{j=1}^C V_c \| L_c-L_j \|^2, \nonumber
\end{align}
where we denote $L_{2D} \left(P\left(\mathbf{X}_t,c\right)|\mathcal{I}_c\right)$ as $L_c$, and $V(\mathbf{X}_t, c)$ as $V_c$ by an abuse of notation. The view-pooling operation finds the best view among the visible cameras that is consistent with other view predictions (the weighted median of $\{L_c\}_{c\in \mathcal{C}}$). 

The view-pooling operation is based on our conjecture that among many views, there exist a few views that can confidently predict an object label. It is robust to noisy recognition outputs as shown in Figure~\ref{Fig:detection} where many false positive bounding boxes are detected. The visibility based confidence measure can suppress inconsistent detection across views, and weighted median pooling can prevent from a view biased $L_{3D}$. This allows the pooled $L_{2D}$ temporally consistent, which makes averaging over time meaningful.

Figure~\ref{Fig:pooling} illustrates the view-pooling operation over all cameras. A set of $L_c$ (bar graphs) at the projected locations $\{P(\mathbf{X},c)\}_{c\in \mathcal{C}}$ are used for the view-pooling that finds the $L_{c^\ast}$ that best represents the distribution of $L_c$. For an illustrative purpose, we highlight the cameras that have high visibility with magenta color, i.e., $V(\mathbf{X},c) > \epsilon_e$. 



\subsection{3D Trajectory Affinity} \label{Sec:aff}
An object that undergoes locally rigid motion provides a spatial cue to identify the affinity between fragmented trajectories. Consider two trajectories $\mathcal{X}_i$ and $\mathcal{X}_j$ that have overlapping lifetime, $\emptyset\neq \mathcal{S} = [T_e^i,T_d^i] \cap [T_e^j,T_d^j]$ where the superscript in $T_e$ and $T_d$ indicates the index of the trajectory. We measure the affinity of the trajectories as follow:
\begin{align}
	A(i, j) = \exp\left(-\left(\|\mathbf{e}_i^j\|/\tau\right)^2\right) \label{Eq:affinity}
\end{align}
where $A \in \mathds{R}^{M\times M}$ is an affinity matrix whose $(i,j)$ entry measures the reconstruction error:
\begin{align}
	\mathbf{e}_i^j = \underset{t-1,t \in  \mathcal{S}}{\operatorname{max}}~~\left\|\mathbf{X}^j_t - \mathbf{R}^i_t\mathbf{X}^j_{t-1} - \mathbf{t}^i_t\right\|.\nonumber
\end{align}
$\mathbf{e}_i^j$ is the Euclidean distance between $\mathbf{X}_t^j$ and the predicted point by its emerging location $\mathbf{X}_{T_e}^j$ via its local transformation $(\mathbf{R}_t^i, \mathbf{t}_t^i) \in SE(3)$ (rotation and translation) learned by the $i^{\rm th}$ trajectory $\mathcal{X}_i$. This measure can be applied to long range trajectories, which establish a strong connection across an object, e.g., left hand to left elbow trajectories. 
$i,j \in \mathcal{T} = \{1,\cdots,M\}$ where $M$ is the number of trajectories. 
Unlike difference of pairwise point distance measure that has been used for trajectory clustering~\cite{joo_iccv_2015}, our affinity takes into account general Euclidean transformation ($SE(3)$) that directly measures rigidity. 

We learn the local transformation $(\mathbf{R}_t^i, \mathbf{t}_t^i)$ of the $i^{\rm th}$ trajectory at each time instant, given a set of neighbors:
\begin{align}
	\mathbf{R}_t^i &= \Delta \mathbf{X}_{t}^{\mathcal{N}_i} \left(\Delta \mathbf{X}_{t-1}^{\mathcal{N}_i}\right)^{-1},~~~~
	\mathbf{t}_t^i &= \mathbf{R}_t^i \mathbf{X}_{t-1}^i - \mathbf{X}_{t}^i \label{Eq:transform} 
\end{align}
where $\Delta \mathbf{X}_{t}^{\mathcal{N}_i}$ is a matrix whose columns are made of relative displacement vectors of neighboring trajectories with respect to $\mathcal{X}_i$, i.e., $\Delta \mathbf{X}_{t}^j = \mathbf{X}^j_t-\mathbf{X}^i_t$ where $j \in \mathcal{N}_i$ is the index of neighboring trajectories. The set of neighbors are chosen as 
\begin{align}
	\mathcal{N}_i = \left\{j\left|\underset{t \in \mathcal{S}}{\operatorname{max}}~~\left\|\mathbf{X}_t^j-\mathbf{X}_t^i\right\| < \epsilon\right.\right\}, \nonumber
\end{align}
where $\epsilon$ is the radius of a 3D Euclidean ball. Note that not all $\epsilon$-neighbors belong to the same object which requires to evaluate the trajectory with Equation~(\ref{Eq:affinity}).  

In practice, evaluating Equation~(\ref{Eq:affinity}) for all trajectories are computationally prohibitive. For example, it requires $10^{10}$ evaluations are needed for 100,000 trajectories\footnote{In our experiments, the number of trajectories is order of $10^4\sim 10^6$.} to fill in all entries in the affinity matrix $A$. Since it is unlikely that far distance trajectories belong to the same object class, we restrict the evaluations only for $\epsilon_a$-neighbors ($\mathcal{N}^a_i$) that are sufficient to cover a large portion of objects and greater $\epsilon$, e.g., $\epsilon_a = 30$cm and $\epsilon = 5$cm. Further, we randomly drop-out connections between neighboring trajectories for computational efficiency. This also increases the robustness of trajectory affinity that is often biased by the density of trajectories. When computing the local transformation in Equation~(\ref{Eq:transform}), we embed RANSAC~\cite{fischler:1981}: choosing random three trajectories from $\epsilon$-neighbors and finding the local transformation that produces the maximum number of inliers.


\begin{figure*}[t]
	\centering  
	\subfigure[System geometry]{\label{Fig:sys}\includegraphics[height=0.125\textheight]{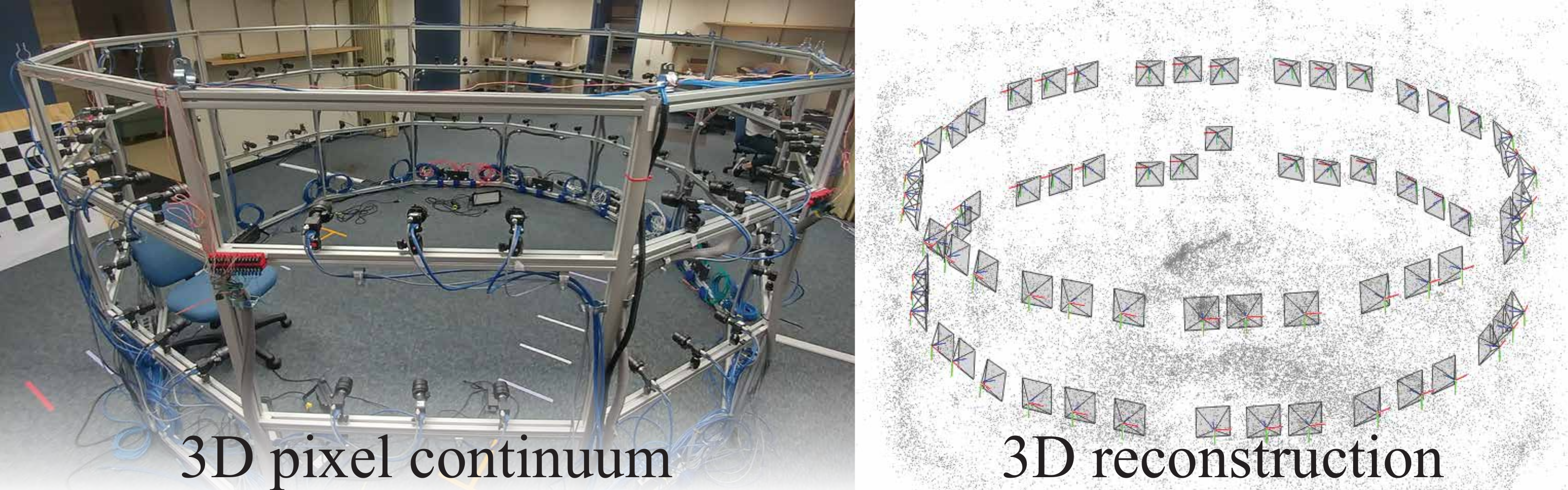}}
	\subfigure[Pixel density]{\label{Fig:pixel_space}\includegraphics[height=0.125\textheight]{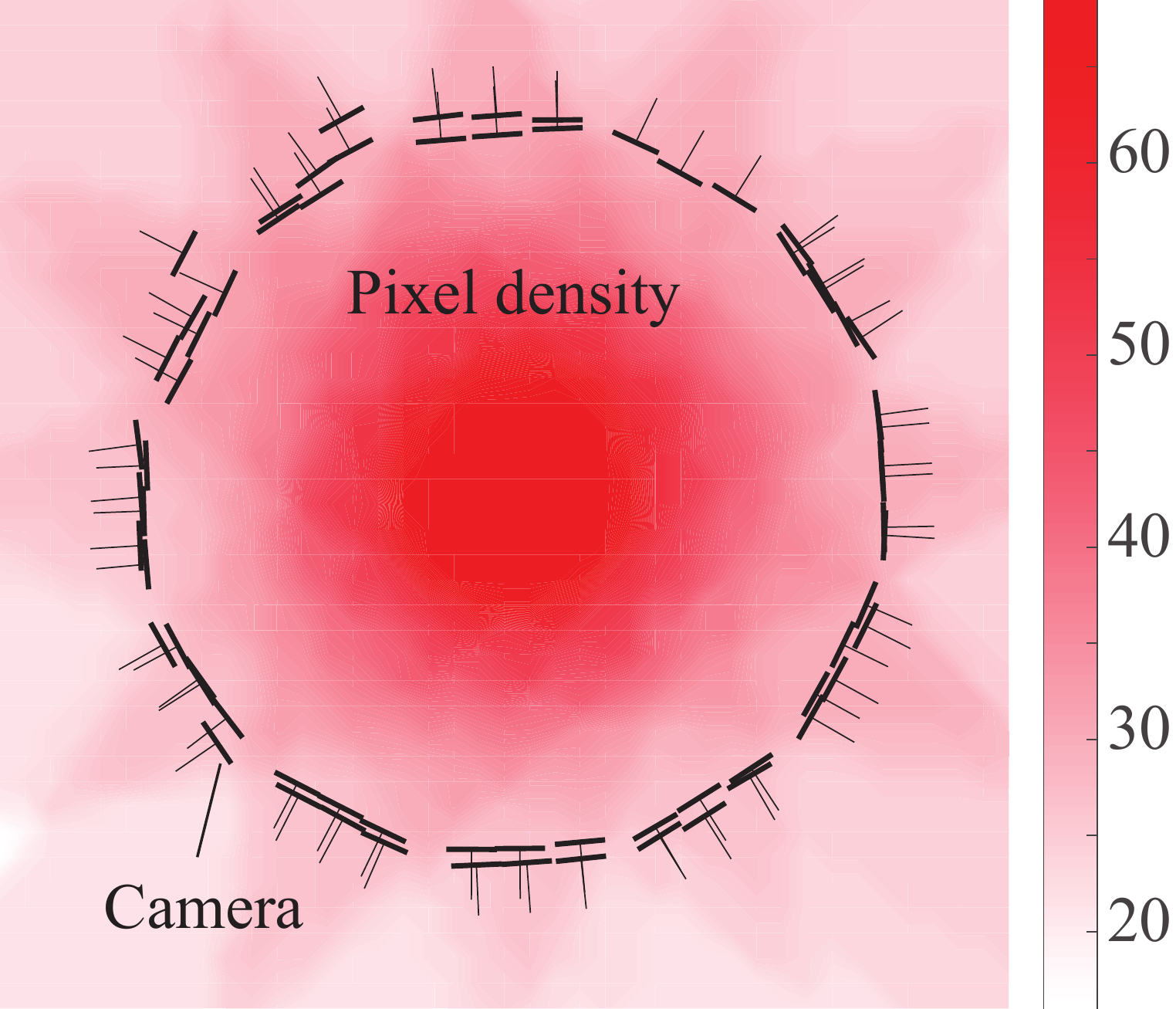}}
	\subfigure[System architecture]{\label{Fig:arch}\includegraphics[height=0.125\textheight]{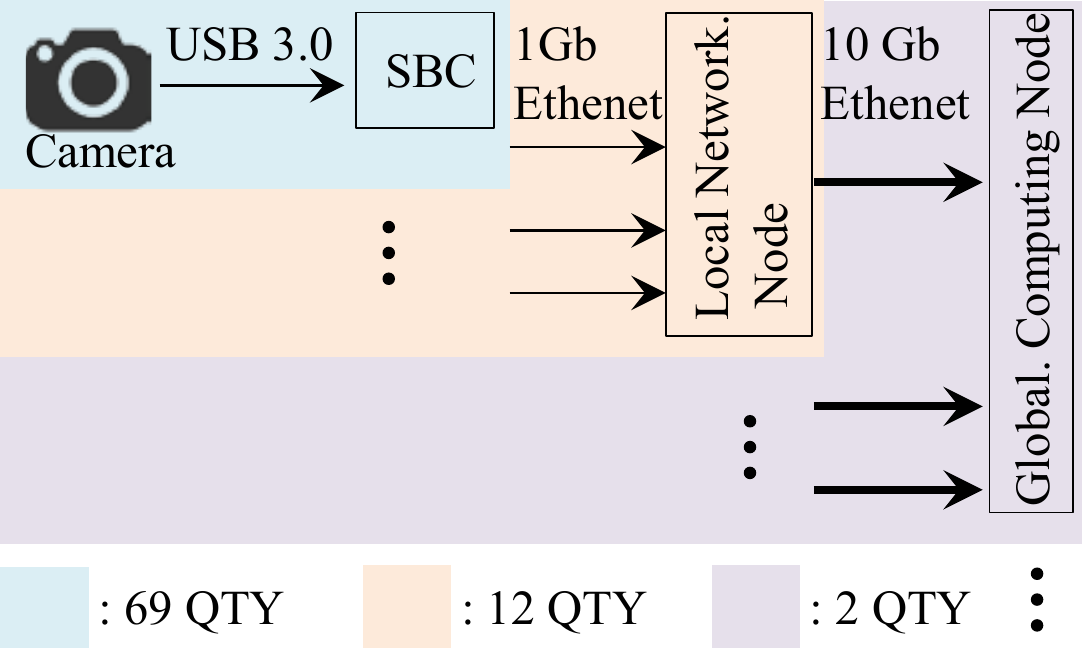}}
	
	\caption{(a) We build a multicamera system composed of 69 cameras running at 30 Hz. (b) The multicamera system creates the 3D pixel continuum where all 3D points in the enclosed space are measured by multiple images. We visualize the pixel density using maximum intensity projection seen from top view. At the center of the stage, more than 60 pixels can measure a unit cm$^3$ cubic. (c) The system architecture is designed using modular units, which makes the system highly scalable. } 
	\label{Fig:system}
\end{figure*}

\subsection{Trajectory Label Inference}
Inspired by multi-class pixel labeling using $\alpha$-expansion~\cite{boykov:2001}, we infer the trajectory labels $U:\mathcal{T}\rightarrow \mathcal{L}$ where $\mathcal{L} =\{1,\cdots,N\}$ is the index set of object classes, by minimizing the following cost:
\begin{align}
	C(U) = \sum_{i\in\mathcal{T}} \phi(l_i, U(i)) + \lambda \sum_{i\in \mathcal{T}} \sum_{j\in\mathcal{N}^a_i} \psi(U(i),U(j)) \label{Eq:cost}
\end{align}
where $\lambda$ is a hyper-parameter that control the weight between data $\phi$ and smoothness $\psi$ costs. 

The data cost can be written as:
\begin{align}
	\phi(l_i, U(i)) &= \left\{\begin{array}{ll}0 &{\rm if}~~ l_i = U(i)\\L_{3D}\left(\mathcal{X}_i\right)_{l_i} & {\rm if}~~l_i \neq U(i)\\\end{array}\right.,\nonumber
\end{align}
where it penalizes the discrepancy between the 3D semantic map predicted by a series of 2D recognitions and assigned label. $L_{3D}\left(\mathcal{X}_i\right)_{l_i}$ is the $l_i^{\rm th}$ entry of $L_{3D}$ that measures the likelihood of $\mathcal{X}_i$ being class $l_i$. 

The smoothness cost can be described by the trajectory affinity:
\begin{align}
	\psi(U(i), U(j)) &= \left\{\begin{array}{ll}0 & {\rm if}~~U(i) = U(j)\\A(i,j) &{\rm if}~~ U(i) \neq U(j)\\\end{array}\right.,\nonumber
\end{align}
where it penalizes the label difference between trajectories that undergo the same local rigid transformation. $l_i$ is the label index computed from $L_{3D}$:
\begin{align}
	l_i = \underset{l \in \mathcal{L}}{\operatorname{argmax}}~~L_{3D} \left(\mathcal{X}_i|\{\mathbf{P}_c,\mathcal{I}_c\}_{c\in\mathcal{C}}\right). \nonumber
\end{align}
Due to multi-class labeling, minimization of Equation~(\ref{Eq:cost}) is highly nonlinear while the iterative $\alpha$-expansion algorithm has been shown a strong convergence towards the global minimum~\cite{boykov:2001,delong:2012}.

\section{3D Trajectory Reconstruction} \label{Sec:reconstruction}

We reconstruct 3D trajectory stream by leveraging the multicamera system described in Section~\ref{Sec:continuum}. In this section, we describe the procedure of the 3D trajectory reconstruction algorithm modified from Joo et al.~\cite{joo_cvpr_2014} to produce denser and more accurate trajectories.  \textbf{(1) Camera calibration} We calibrate the intrinsic parameter of each camera (focal length, principal points, and radial lens distortion), independently, and use standard structure from motion to calibrate extrinsic parameters (relative rotation and translation). In the bundle adjustment, the extrinsic and intrinsic parameters are jointly refined. To accelerate further image based matching, we learn the image connectivity graph~\cite{SNAVELY-IJCV08} $\mathcal{G}_m=(\mathcal{V}_m,\mathcal{E}_m)$ through exhaustive pairwise image matching, e.g., two cameras that have more than 90 degree apart are unlikely to match to each other. \textbf{(2) Point cloud triangulation} At each time instant, we find dense feature correspondences using grid-based motion statistics (GMS)~\cite{bian:2017} among $\mathcal{G}_m$ and triangulate each 3D point $\mathbf{X}$ with RANSAC. The initial visibility for the $c^{\rm th}$ camera is set to $V(\mathbf{X}, c) = \exp(-\left(\|P(\mathbf{X},c)\|/\sigma\right)^2)$ where the $\sigma$ is the tolerance of the reprojection error.  \textbf{(3) 3D point tracking} The triangulated points are used for build trajectory stream. For each point $\mathbf{X}_t$ at the $t$ time instant, we project the point onto the visible set of cameras, i.e., $P(\mathbf{X}_t, c\in\mathcal{V})$ where $\mathcal{V} = \{j|V(\mathbf{X}_{t-1},c) > \epsilon_s\}$ where $\epsilon_s$ is the threshold for the probability of visibility. These projected points are tracked in 2D using optical flow and triangulated with RANSAC to form $\mathbf{X}_{t+1}$. Similar to the visibility initialization, the probability of visibility $V(\mathbf{X}_{t+1}, c)$ is updated using reprojection error. We iterate this process (tracking$\rightarrow$triangulation$\rightarrow$visibility update) until the average reprojection is higher than 2 pixels or the number of visible cameras $|\mathcal{V}|$ is less than 2.

\section{3D Pixel Continuum Design} \label{Sec:continuum}

To demonstrate the 3D pixel continuum where every 3D point is observed by multiple cameras, we build a large scale multicamera system composed of 69 cameras as shown in Figure~\ref{Fig:sys}. Two rows of the cameras enclose cylindrical space (3m diameter $\times$ 2.5m height) that facilitates capturing diverse human interactions. A camera produces a HD resolution image (1280$\times$1024) where the maximum pixel density per unit cm$^3$ reaches to more than 60 pixels. It runs at 30 Hz precisely triggered by a master camera node: the master camera sends PWM signal through General Purpose Input/Output (GPIO) port when its shutter opens, which triggers  the rest 68 slave cameras, achieving sub-nano second accuracy. To alleviate the trigger signal attenuation due to a number of camera connections, we design a signal amplifier that can feed the targeted electric current.

All cameras produce a shear amount of visual data at each second (280 GB/s), which introduces severe data traffic in the global computing node. Instead, we modularize the image processing using a single board computer (SBC): the image data stream from each camera is transferred through USB 3.0 to its own SBC that is dedicated to JPEG image compression, resulting in $\sim 400$ KB/image with minimal loss of image quality. This compressed data is transferred to two global computing nodes through multiples of 10 Gb Ethernet network switches. The global computing nodes write the data into designated PCIe interfaced solid state drives (SSD). The architecture is summarized in Figure~\ref{Fig:arch}.

The key features of the system design is scalability and cost effectiveness. The modularized system design allows increasing the number of cameras and size of the system without introducing system complexity: the module of camera-SBC-Network switch can be augmented in the existing system. Also the hardware frame is build on modular T-slotted aluminum frame where the modification of geometric camera placement can be easily customizable. All parts including hardware, electronic devices, and cameras are commodity items where no system specific design is needed.








\begin{figure*}[t]
	\centering  
	\subfigure[Pet int.]{\includegraphics[height=0.152\textheight]{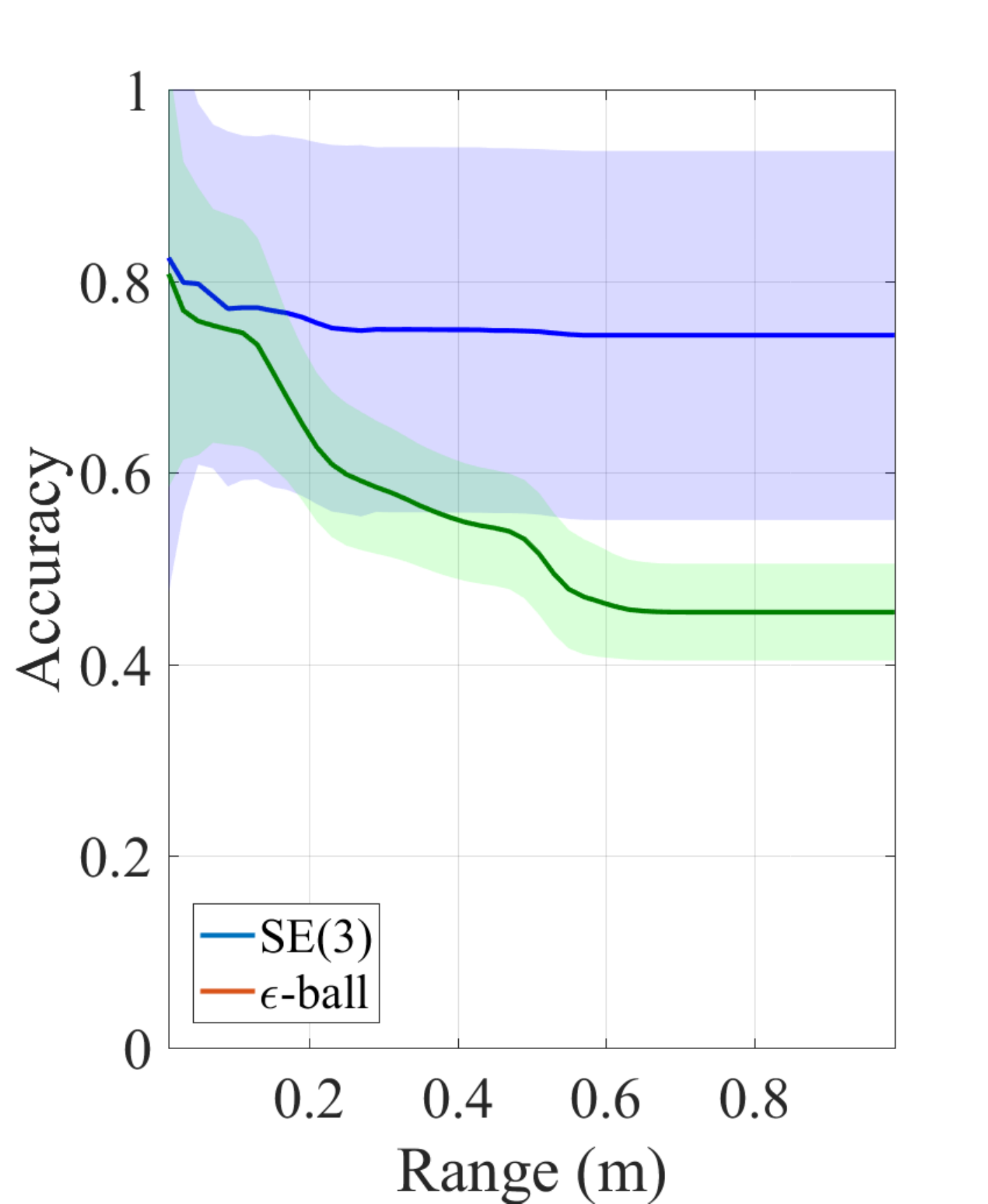}}
	\subfigure[L. dance]{\includegraphics[height=0.152\textheight]{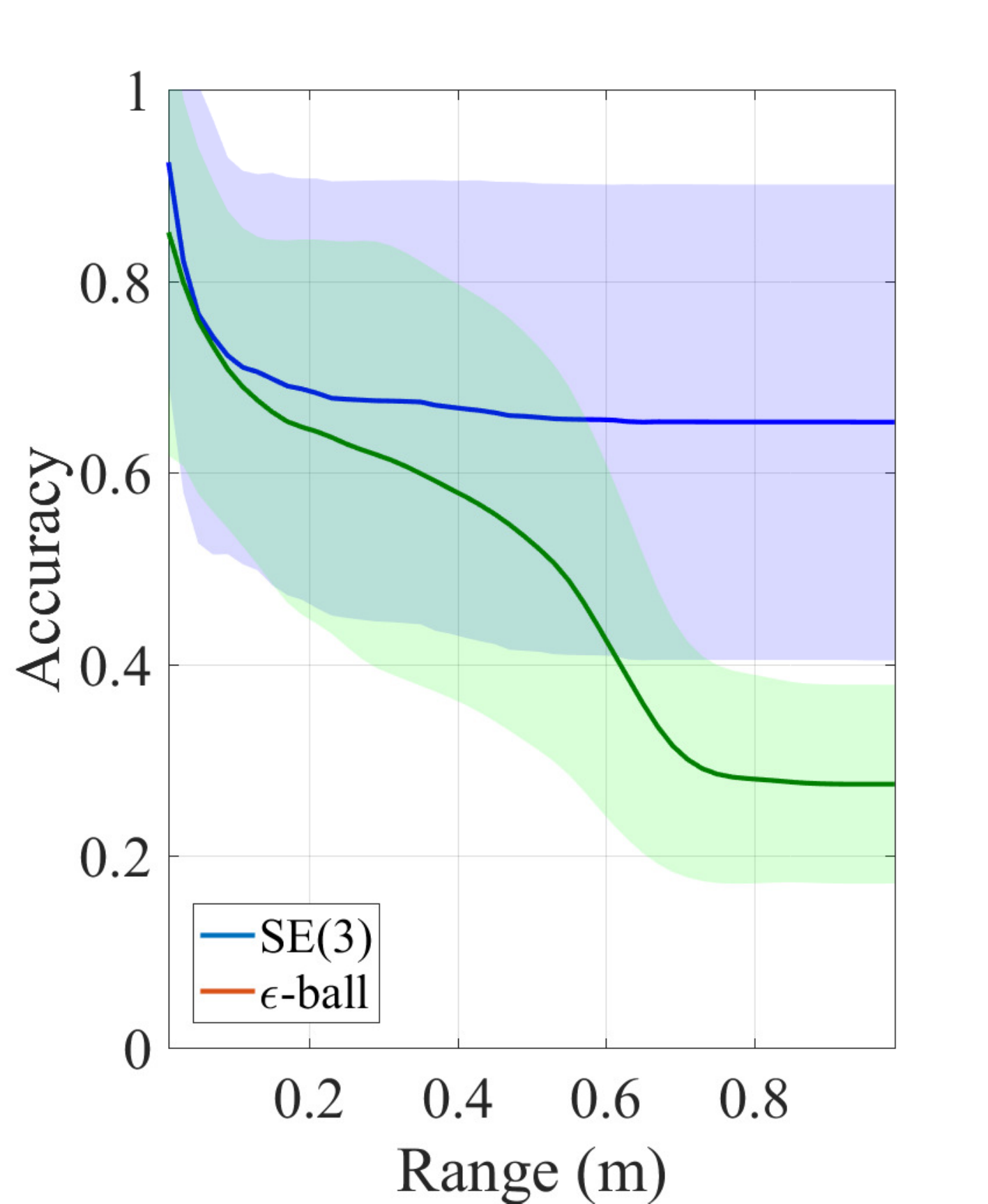}}
	\subfigure[K-Pop]{\includegraphics[height=0.152\textheight]{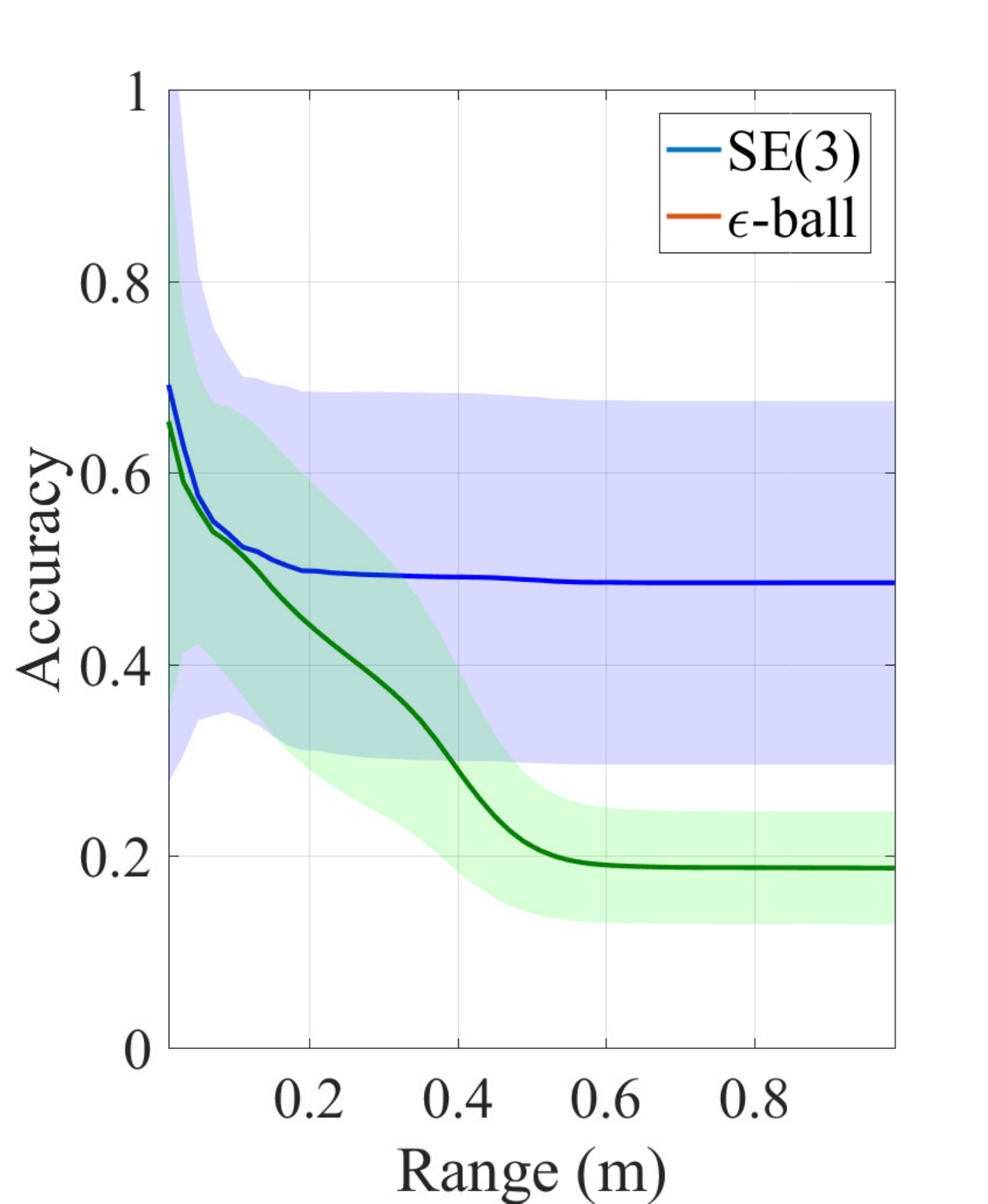}}
	\subfigure[Bike]{\includegraphics[height=0.152\textheight]{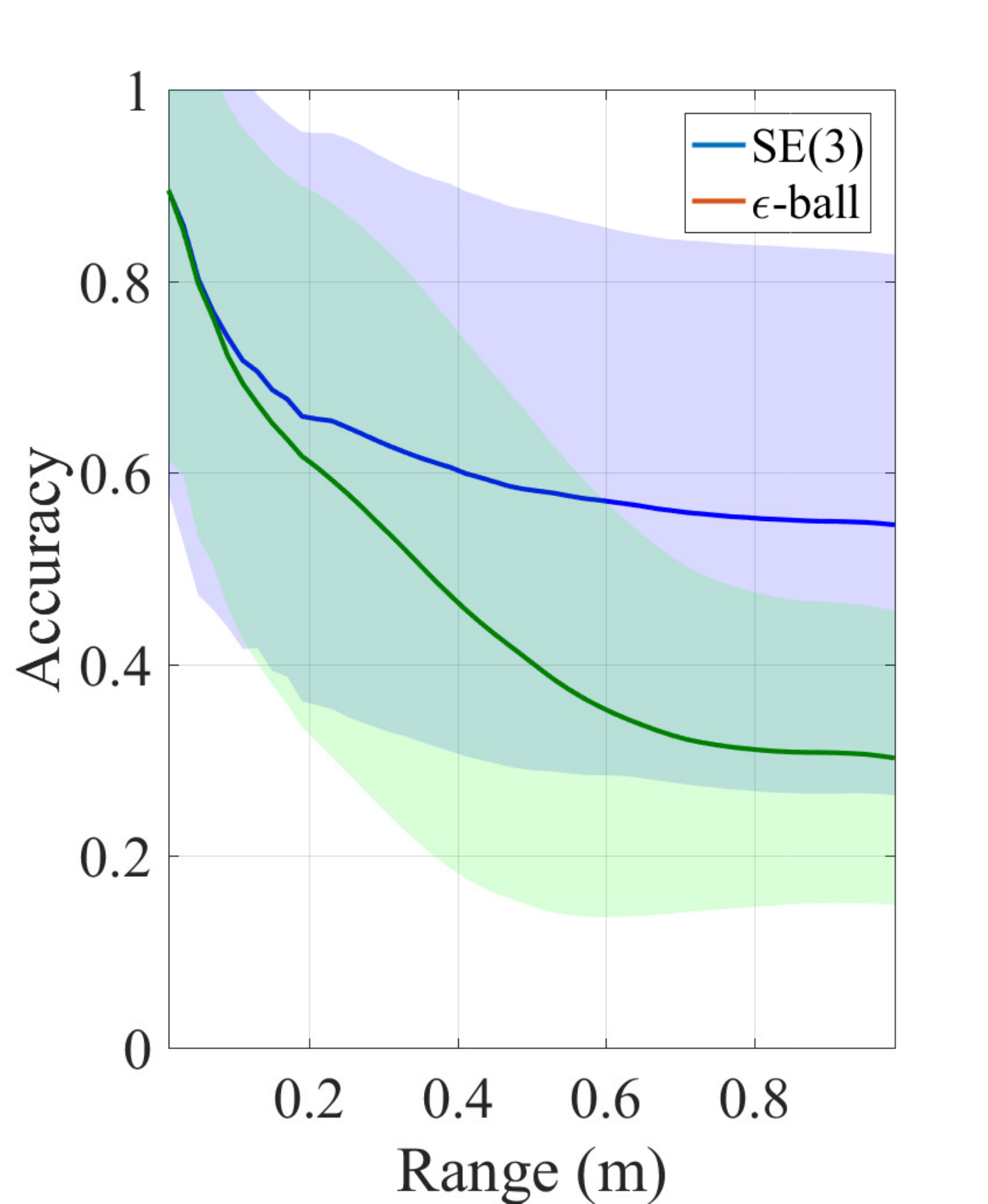}}
	\subfigure[Tennis]{\includegraphics[height=0.152\textheight]{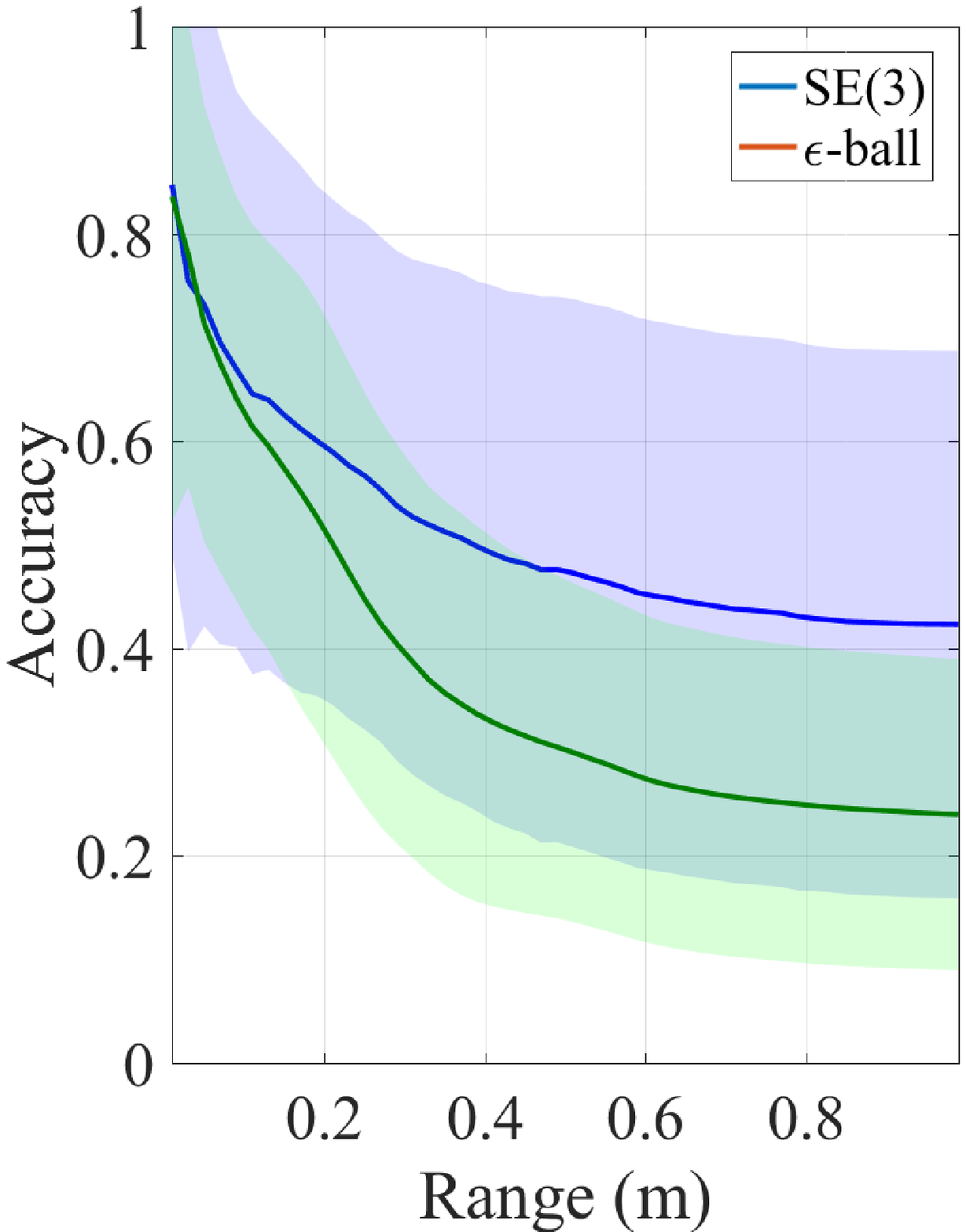}}
	\subfigure[Basketball II]{\includegraphics[height=0.152\textheight]{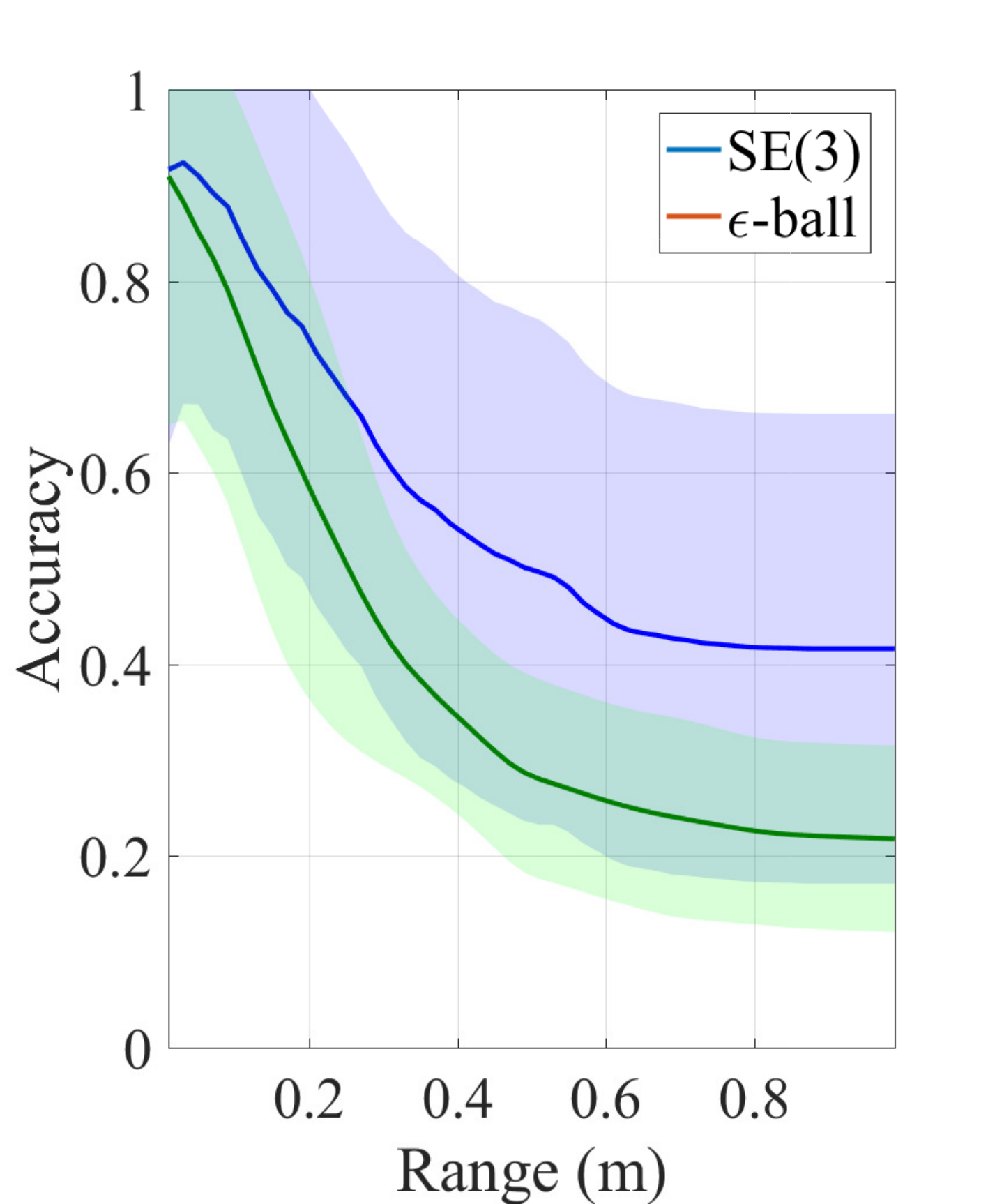}}
	\caption{We evaluate the effectiveness of our affinity map computed by estimating local Euclidean transformation SE(3). While the effectiveness of $\epsilon_s$-neighbors diminishes rapidly after 10 cm, our method still holds for longer range, e.g., 1 m.} 
	\label{Fig:aff}
\end{figure*}

\section{Results}
To validate our semantic trajectory reconstruction algorithm, we evaluate on real-world datasets collected by the 3D pixel continuum described in Section~\ref{Sec:continuum}. 

\subsection{Human Interaction Dataset}
9 new vignettes that include diverse human interactions are captured: \textbf{Pet interaction}: A dog owner naturally interacts with her dog: ask him to sit, turn around and jump. The dog also plays with his doll and seek snack while walking around with the owner. This pet interaction demonstrates strength of our system, i.e., reconstructing fine detailed interactions, not limited to humans~\cite{joo_iccv_2015}; \textbf{International Latin ballroom dance}: Two sport dancers practice for Cha-cha style dance competition where the physical interactions between them are highly stylized. The dancers wear textureless black suit and skirt where semantic labeling is likely noisy; \textbf{K-Pop group dance}: Two experienced K-Pop dancers perform the group break dance. The dances are designed to be synchronized, jerky, and fast; \textbf{Object manipulation}: Two students manipulate various objects such as doll, flowerpot, monitor, umbrella, and hair drier in a cluttered environments. This vignette demonstrates that the system is able to handle multiple objects; \textbf{Bicycle riding}: A person rides a bicycle that induces large displacement. This interaction introduces significant occlusion, i.e., the person is a part of the bicycle; \textbf{Tennis swing}: A person practices fore- and back-hand strokes with a tennis racket. The tennis racket is often difficult to detect as the racket head is mostly transparent; \textbf{Basketball I}: A student player practices dribbling which includes fast ball motion; \textbf{Basketball II}: An other player tries to block the opponent's motion that includes severe occlusion between players. 

\begin{figure*}[t]
	\centering  
	\subfigure[Pet int.]{\includegraphics[height=0.152\textheight]{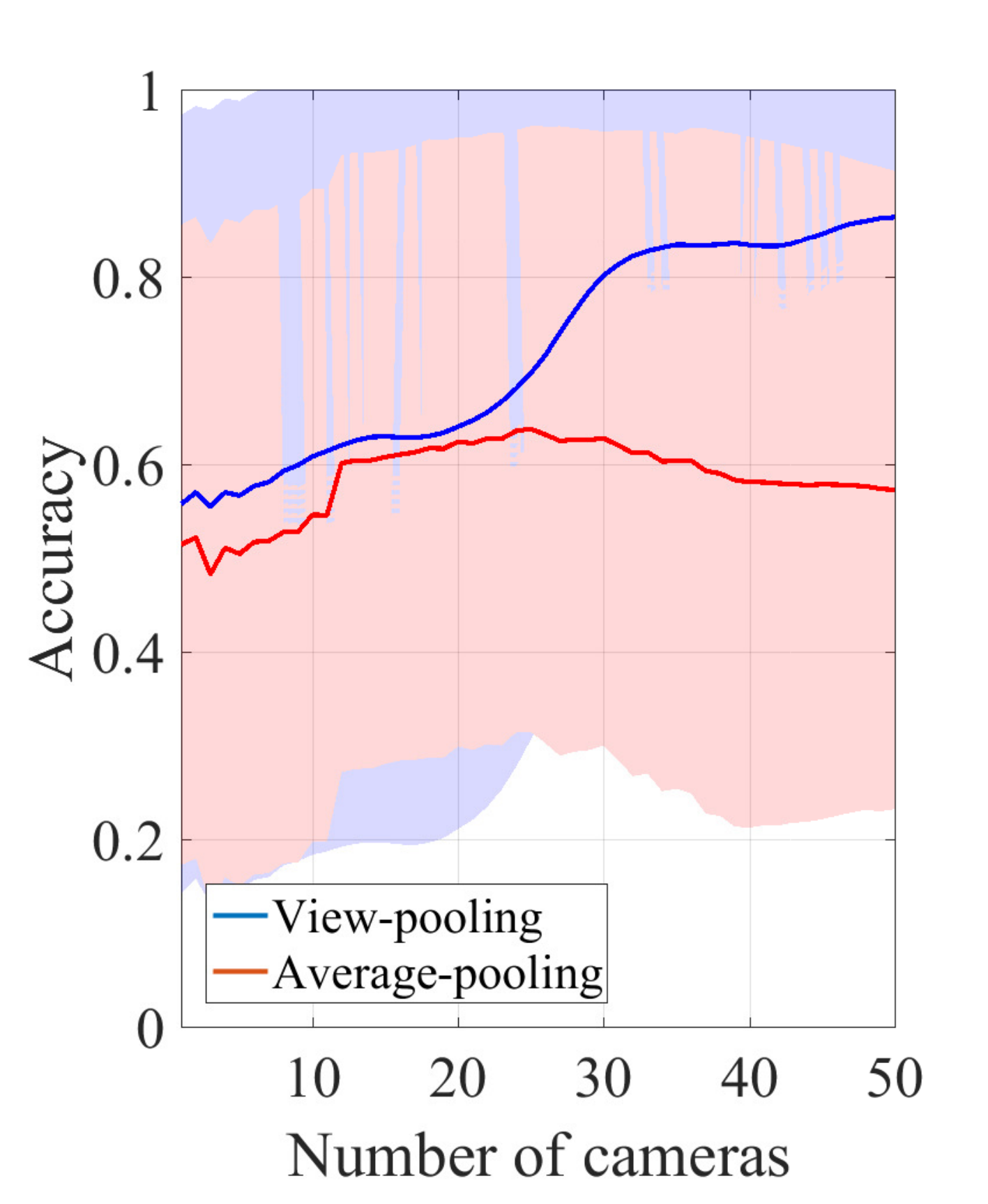}}
	\subfigure[L. dance]{\includegraphics[height=0.152\textheight]{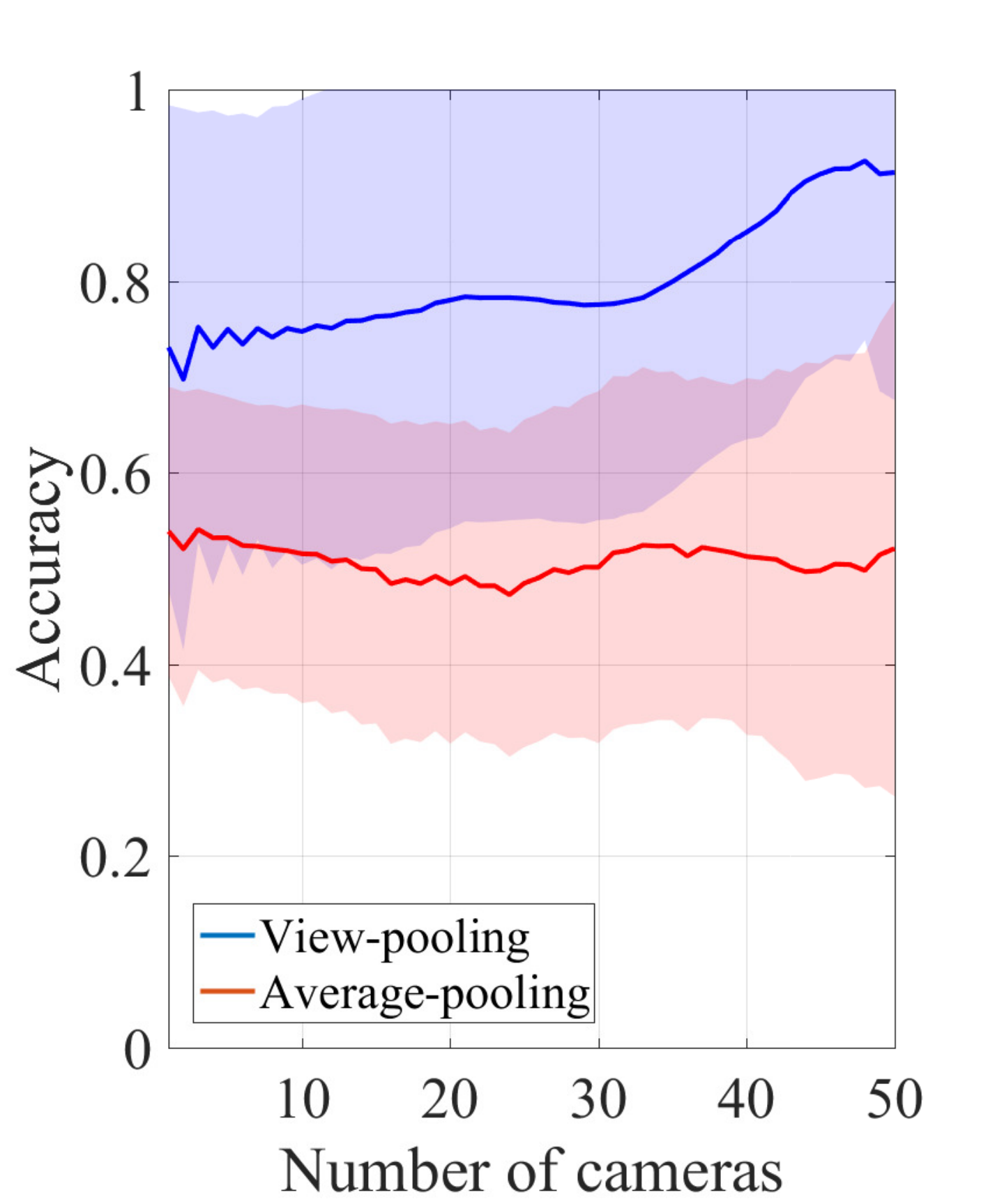}}
	\subfigure[K-Pop]{\includegraphics[height=0.152\textheight]{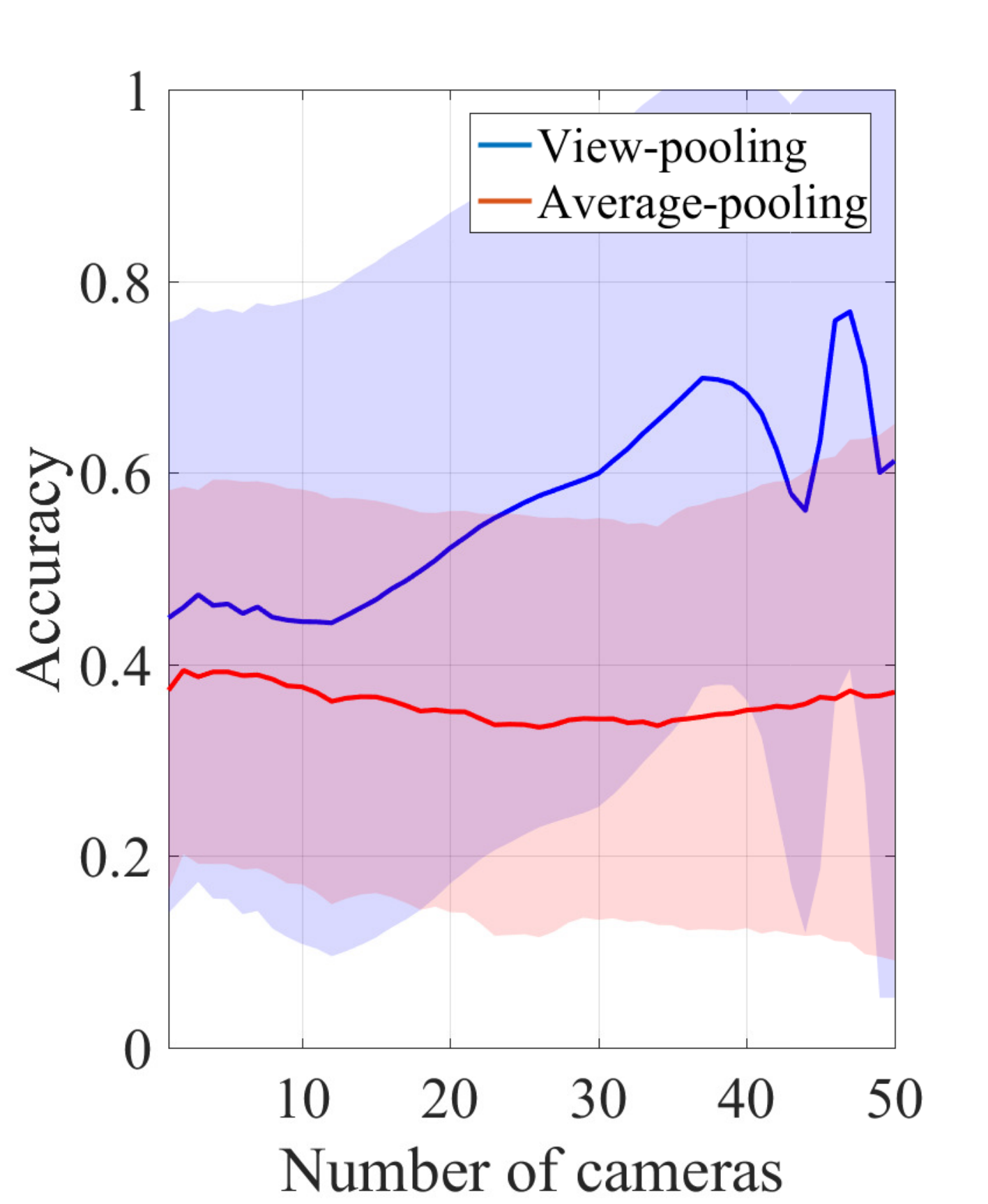}}
	\subfigure[Tennis]{\includegraphics[height=0.152\textheight]{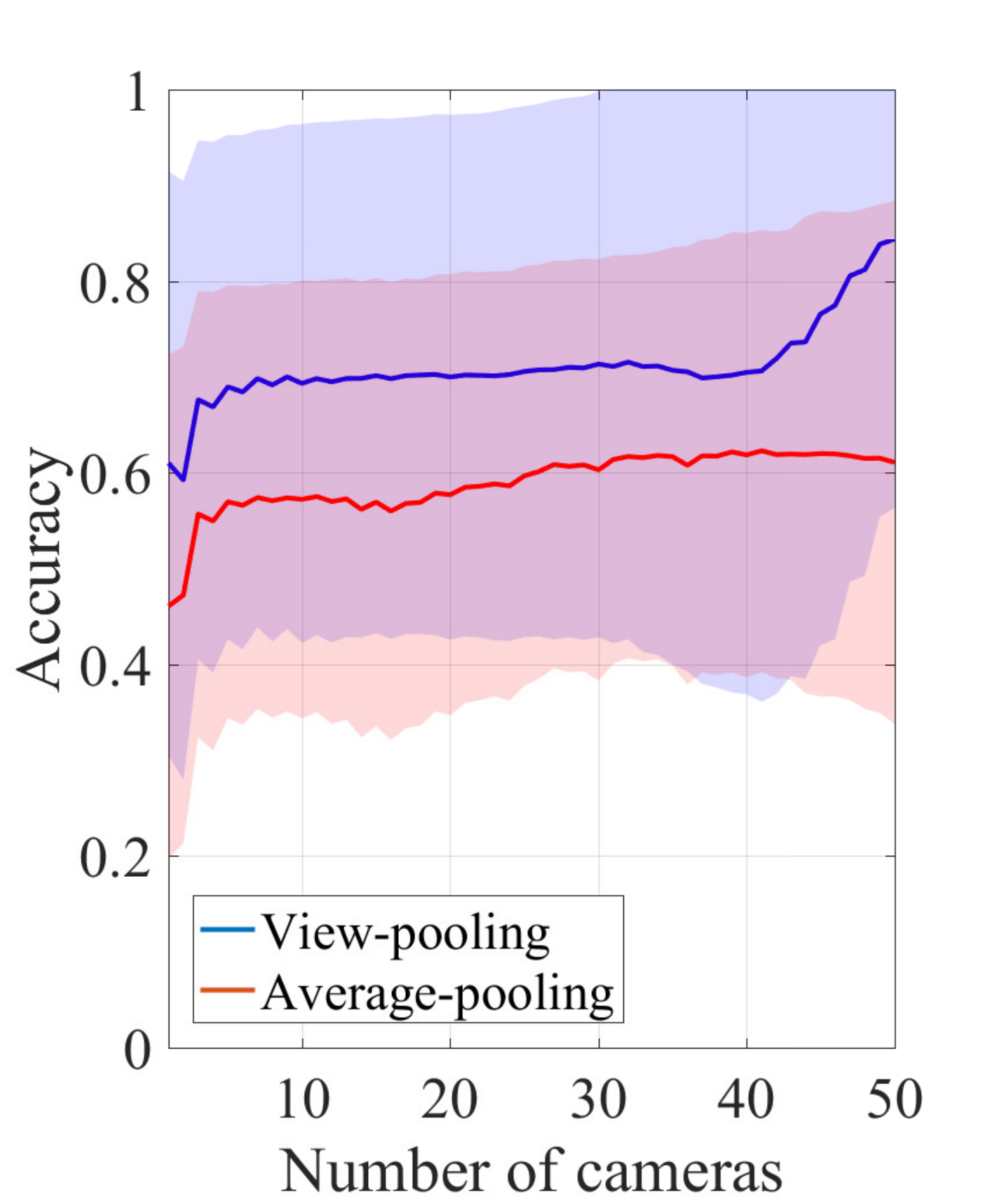}}
	\subfigure[Basketball I]{\includegraphics[height=0.152\textheight]{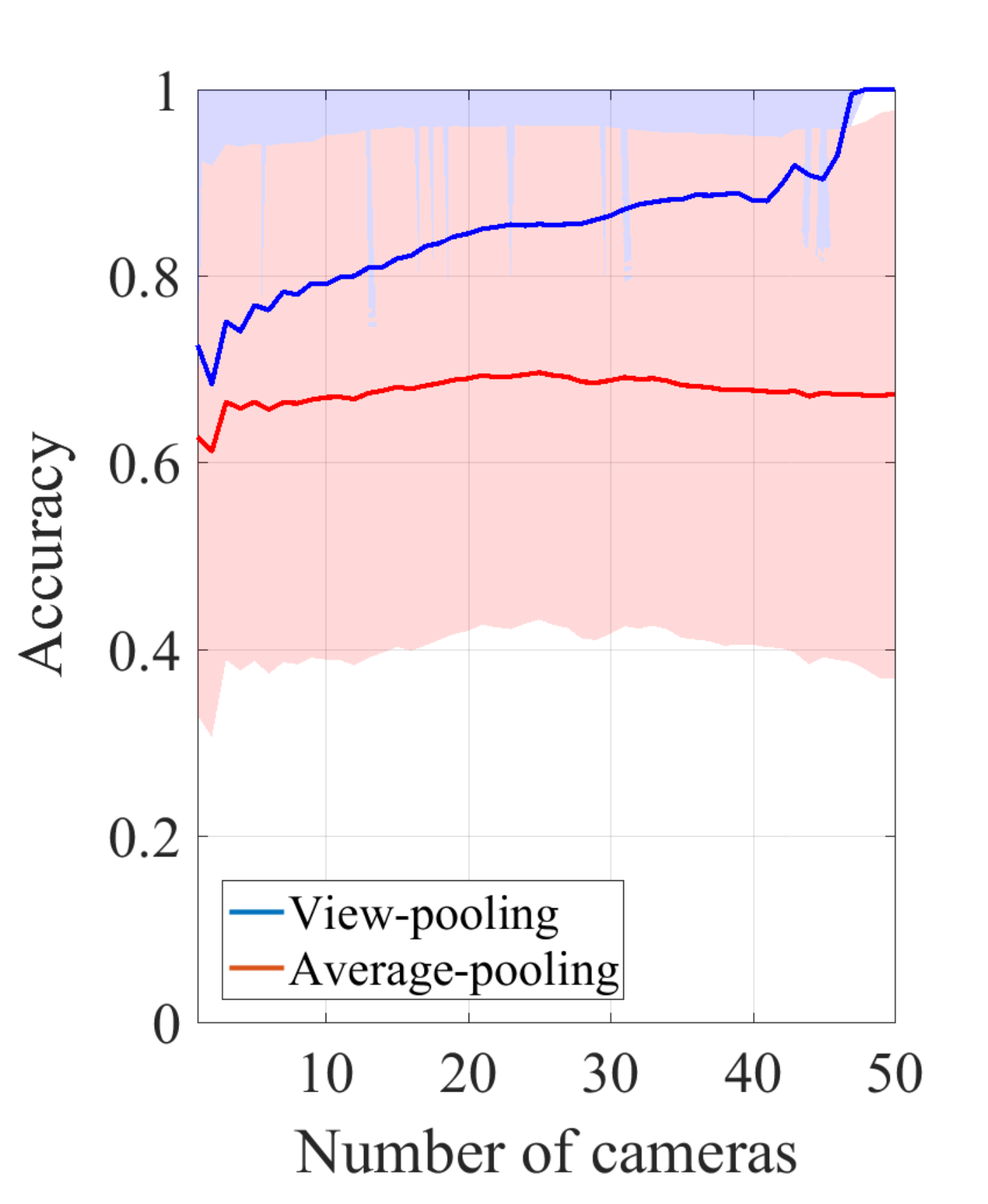}}
	\subfigure[Basketball II]{\includegraphics[height=0.152\textheight]{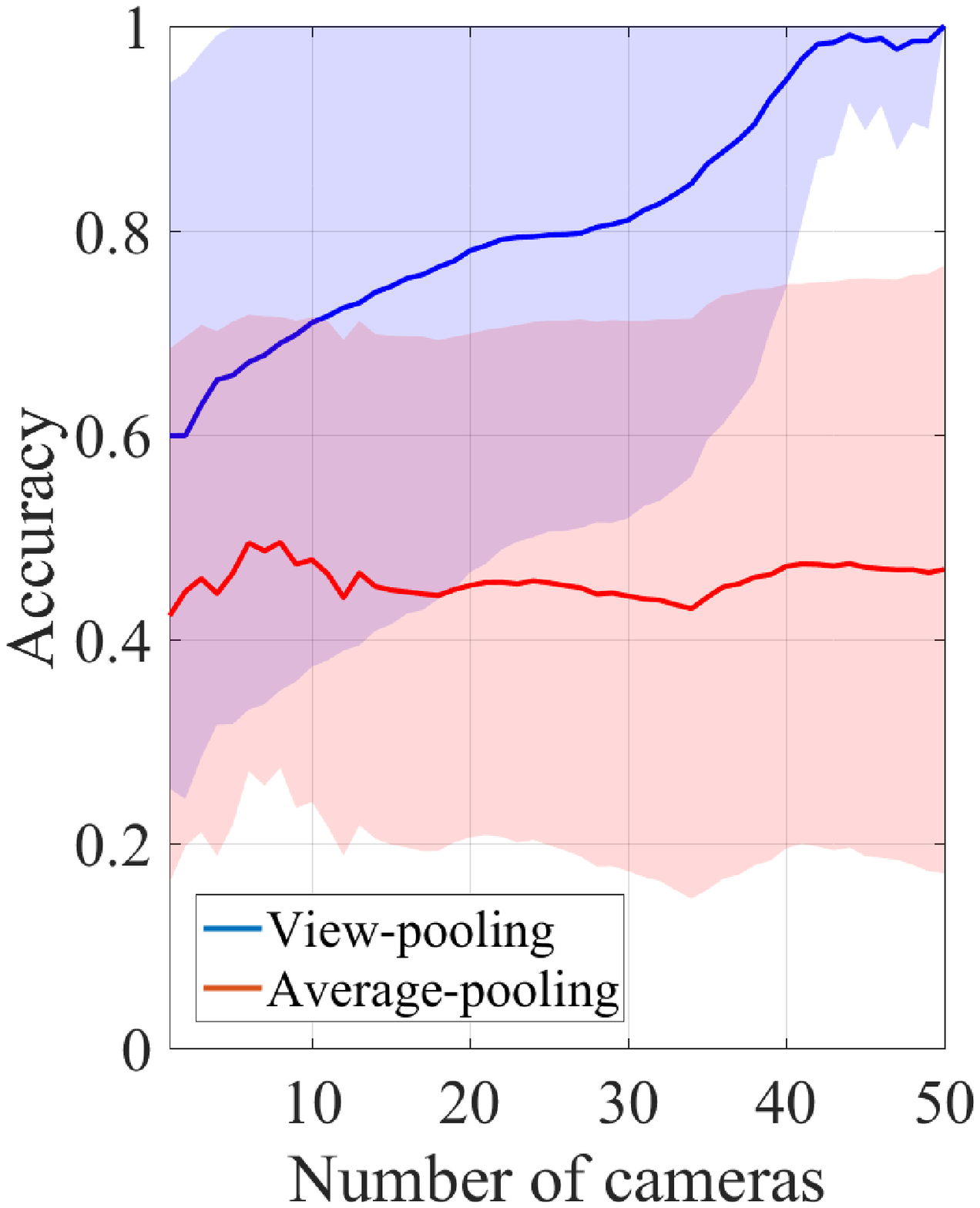}}
	\caption{We evaluate semantic label prediction via an ablation study: to use a subset of cameras to assign the semantic labels to the trajectories and validate the labels by comparing the labels of projections with the held-out images. Our view-pooling method outperforms the average-pooling with large margin for all sequences. } 
	\label{Fig:pre}
\end{figure*}

\subsection{Quantitative Evaluation}
We quantitatively evaluate our representation and algorithm in terms of three criteria: (1) robustness of 3D semantic map (view-pooling); (2) effectiveness of the affinity measure; and (3) predictive validity of semantic labels where all datasets are used for the evaluations. Note that as no ground truth data or benchmark dataset is available, we conduct ablation studies to validate our methods. 

\noindent\textbf{Robustness of 3D semantic map} We introduce the view-pooling operation that takes the weighted median of recognition confidence based on visibility. This operation allows robustly predicting the 3D semantic map $L_{3D}$ as it is not sensitive to erroneous detection. To evaluate its robustness, we measure the temporal consistency of the view-pooling operation along a trajectory. Ideally, the view-pooled recognition confidence should remain constant across time as it belongs to the trajectory of the same object. We compare the view-pooling with average-pooling across randomly all cameras using normalized correlation measure across time, i.e., $NC(L_{\rm vp}^0,L_{\rm vp}^t)$ where $L_{\rm vp}^t$ is the view-pooled recognition confidence at the $t$ time instant. We summarize the results on all sequences in Table~\ref{table:robust}. Our method shows a graceful degradation as time progress up to 15 seconds while the average-pooling is highly biased by noisy recognition, which produces drastic performance gradation (no temporal coherence).  

\begin{table}[h]
	\centering
	\scriptsize
	\begin{tabular}{l|c|c|c|c}
		\hline
		Time (second) & 1s & 3s & 5s & 7s\\
		\hline
		View pool & \textbf{0.96}$\pm$0.01 & \textbf{0.90}$\pm$0.02 & \textbf{0.89}$\pm$0.03  & \textbf{0.88}$\pm$0.02 \\
		Ave. pool & 0.43$\pm$0.10 & 0.44$\pm$0.10 & 0.43$\pm$0.10  & 0.48$\pm$0.09\\ 
		\hline
		Time (second) & 9s & 11s & 13s & 15s\\
		\hline
		View pool & \textbf{0.89}$\pm$0.02 & \textbf{0.88}$\pm$0.03 & \textbf{0.87}$\pm$0.05 & \textbf{0.79}$\pm$0.08\\
		Ave. pool & 0.44$\pm$0.09 & 0.43$\pm$0.10 & 0.42$\pm$0.10 & 0.37$\pm$0.10\\
		\hline
	\end{tabular}
	\caption[Time consistency of 3D semantic map]{Time consistency of 3D semantic map}
	\label{table:robust}
\end{table}

\noindent\textbf{Effectiveness of affinity measure} We compute the affinity based on local transformation per trajectory. This method is highly effective to relate with long term fragmented trajectories. We compare the validity of our affinity measure with that of $\epsilon_s$-neighbors ($\mathcal{N}_s$), i.e., the distance between trajectories over time remains less than $\epsilon_s$. To evaluate, two neighboring trajectories for both methods are randomly chosen and projected onto cameras. Concretely, we measure $\sum_{j\in \mathcal{N}_s} E(i,j)$ where 
\begin{align}
	E(i,j) = \left\{\begin{array}{ll}0 & {\rm if}~ L(P(\mathbf{X}_t^i,c)|\mathcal{I}_c) = L(P(\mathbf{X}_t^j,c)|\mathcal{I}_c)\\
		1 & {\rm otherwise}\end{array}\right..\nonumber
\end{align}
$L:\mathds{R}^2\rightarrow\mathcal{L}$ outputs the semantic label index given the 2D projection. If the measure is small, it indicates that the neighbors are correctly identified. Figure~\ref{Fig:aff} illustrates the comparison over 6 different sequences. Each one has different global and local motion. If the motion is largely global, the affinity measure can confuse as multibody motion is identified as a rigid body motion as shown in Basketball II. Nonetheless, our method outperforms the $\epsilon_s$-neighbors for all sequences. In particular, it shows much stronger performance at long range trajectories (0.6-1 m), which makes the large scale label inference possible. 





\noindent\textbf{Predictive validity of 3D semantic label} We evaluate the semantic label inference via cross validation scheme. We label a 3D trajectory with a subset of cameras and project onto the held-out camera to evaluate the predictive validity. Ideally, the trajectory label should be consistent with any view as visibility is considered, and therefore, the projected label must agree with the recognition result. As we infer the semantic labels of the trajectories jointly by consolidating multiple view recognition, the number of cameras plays a key role in the inference. We test the predictive validity by changing the number of cameras to label trajectories as shown in Figure~\ref{Fig:pre}. When the number of cameras is few, e.g., 1-5, our method using view-pooling performs similarly with average-pooling. However, the performance quickly is boosted as the number of camera increases, i.e., in most cases, it produces more than 0.6 accuracy at 20 cameras for inference.

\begin{figure}[t]
	\centering  
	\includegraphics[width=0.48\textwidth]{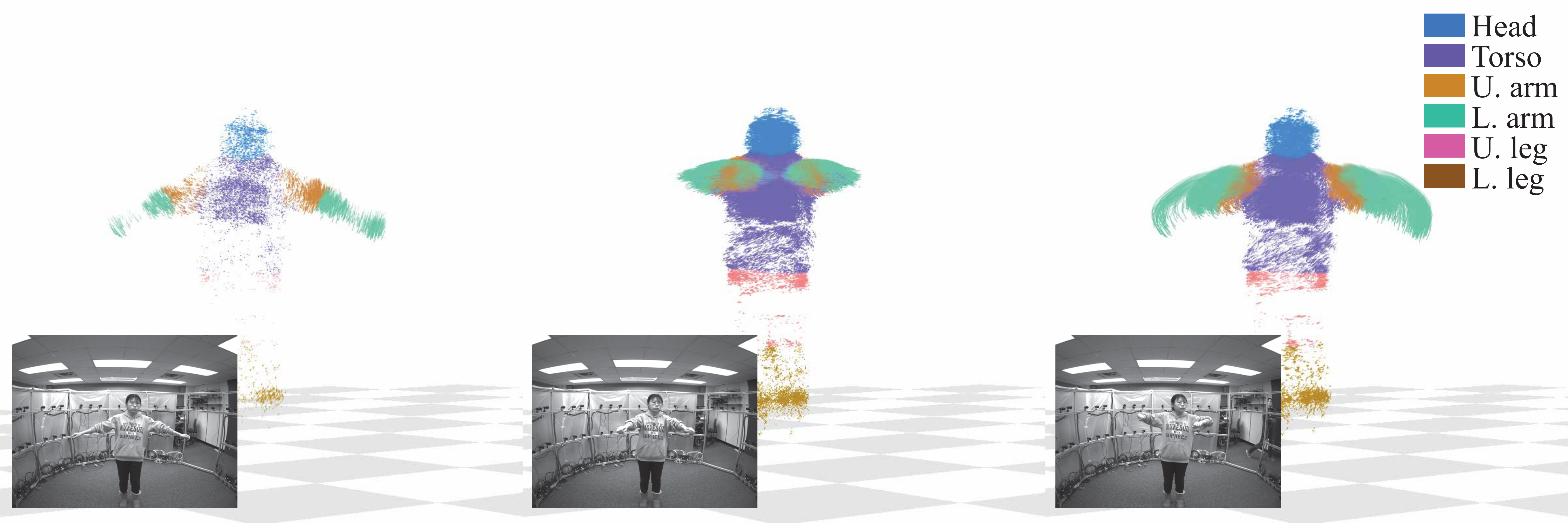}
	\caption{Range of motion} 
	\label{Fig:rom}
\end{figure}

\begin{figure}[t]
	\centering  
	\includegraphics[width=0.48\textwidth]{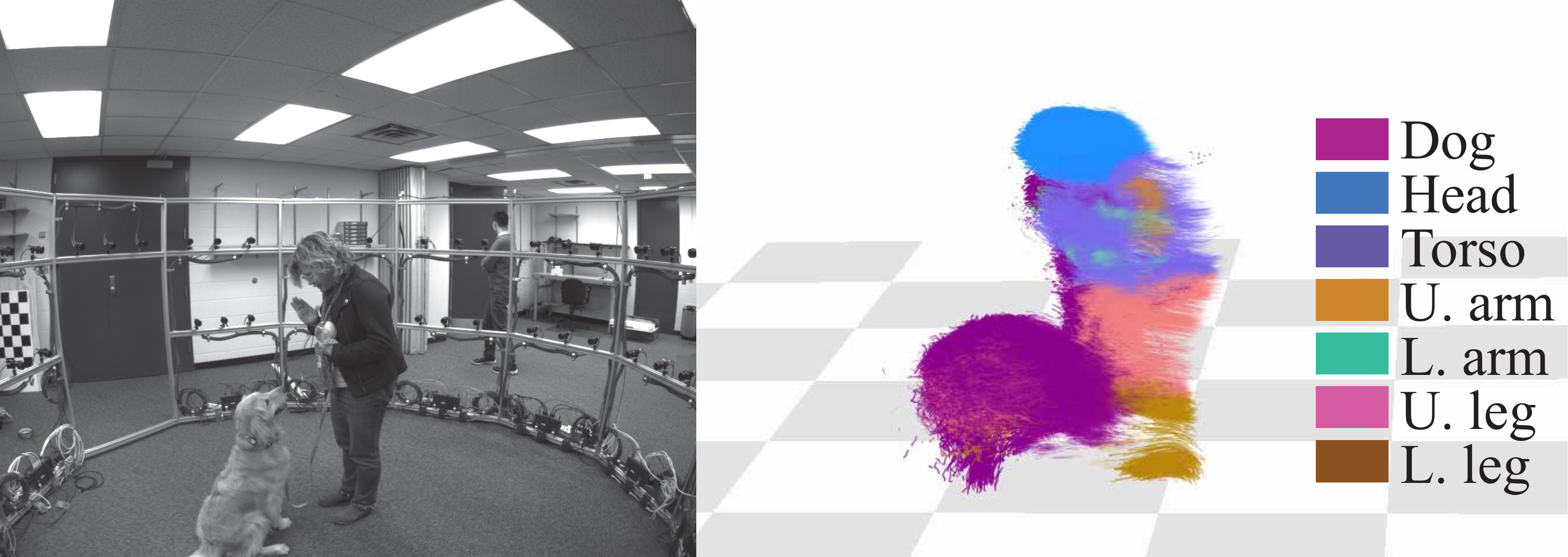}
	\caption{Pet interaction} 
	\label{Fig:dog}
\end{figure}

\begin{figure*}[t]
	\centering  
	\subfigure[Basketball II]{\includegraphics[width=\textwidth]{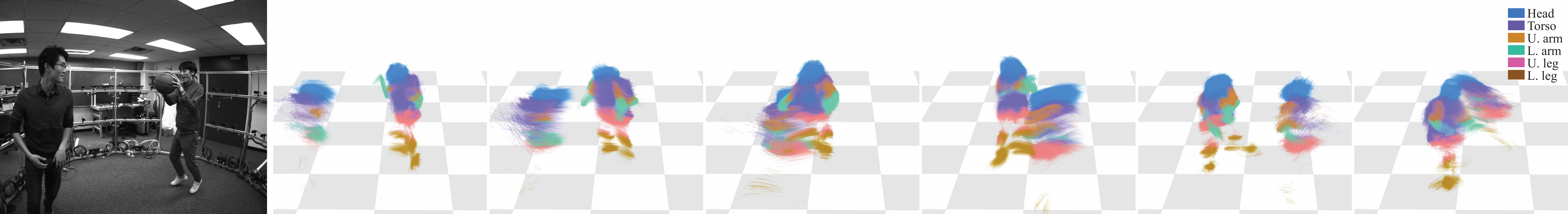}}
	\subfigure[Latin dance]{\includegraphics[width=\textwidth]{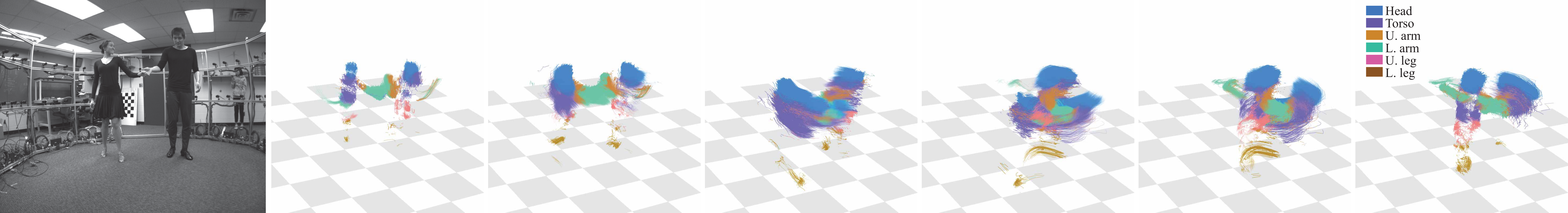}}
	\subfigure[K-Pop]{\includegraphics[width=\textwidth]{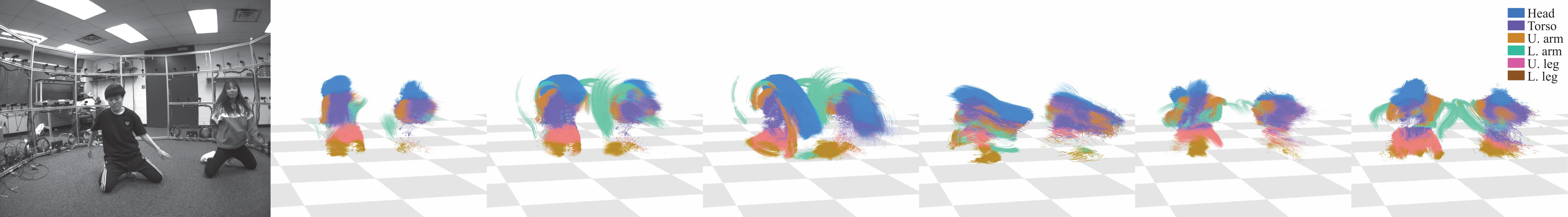}}
	\subfigure[Tennis]{\includegraphics[width=\textwidth]{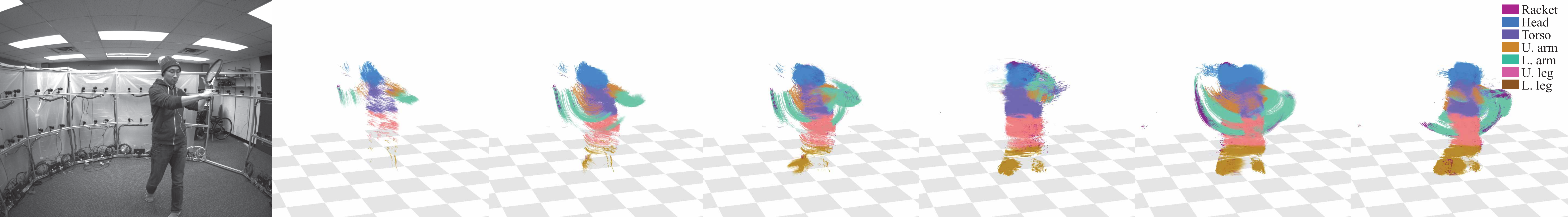}}
	\caption{Qualitative evaluation. Best seen in color. For an illustrative purpose, the last 30 frames of the trajectories are visualized.} 
	\label{Fig:qual}
\end{figure*}

\subsection{Qualitative Evaluation}
We apply our method to reconstruct dense semantic trajectories in 3D as shown in Figure~\ref{Fig:teaser},~\ref{Fig:rom}~\ref{Fig:dog}, and~\ref{Fig:qual}. The colors of the trajectories indicate the semantic labels.  

\section{Discussion}

We present an algorithm to reconstruct semantic trajectories in 3D using a large scale multicamera system. This problem is challenging because of fragmented trajectories and noisy/coarse recognition in 2D. We introduce a new representation to encode the visual semantics to each trajectory called 3D semantic map that allows us to consolidate multiple view noisy recognition results by leveraging view pooling based on their visibility and recognition confidence. 3D spatial relationship between fragmented trajectories is modeled by local rigid transformation that can establish the connection between long range trajectories. These two cues are integrated into a graph-cut formulation to infer precise labeling of the trajectories. Note that Our framework is not specific to the choice of the 2D recognition models. 

The first wave of the optic technology enabled cameras to be emerged and embedded in our space. The second wave will be {\em connectedness}: multiple cameras will measure our interactions and cooperatively understand their semantic meaning. This paper takes the first bold step towards establishing a computational basis for understanding 3D semantics at fine scale.

{\small
	\bibliographystyle{ieee}
	\bibliography{egbib}

\begin{thebibliography}{10}\itemsep=-1pt

\bibitem{akhter:2009}
I.~Akhter, Y.~Sheikh, and S.~Khan.
\newblock In defense of orthonormality constraints for nonrigid structure from
  motion.
\newblock In {\em CVPR}, 2009.

\bibitem{akhter:2008}
I.~Akhter, Y.~Sheikh, S.~Khan, and T.~Kanade.
\newblock Nonrigid structure from motion in trajectory space.
\newblock In {\em NIPS}, 2008.

\bibitem{Akhter:2012:BilinearBasis}
I.~Akhter, T.~Simon, S.~Khan, I.~Matthews, and Y.~Sheikh.
\newblock Bilinear spatiotemporal basis models.
\newblock {\em SIGGRAPH}, 2012.

\bibitem{ambady:1992}
N.~Ambady and R.~Rosenthal.
\newblock Thin slices of expressive behavior as predictors of interpersonal
  consequences: A meta-analysis.
\newblock {\em IJCV}, 1992.

\bibitem{avidan:2000}
S.~Avidan and A.~Shashua.
\newblock Trajectory triangulation: 3{D} reconstruction of moving points from a
  monocular image sequence.
\newblock {\em PAMI}, 2000.

\bibitem{bian:2017}
J.~Bian, W.-Y. Lin, Y.~Matsushita, S.-K. Yeung, T.~D. Nguyen, and M.-M. Cheng.
\newblock Gms: Grid-based motion statistics for fast, ultra-robust feature
  correspondence.
\newblock In {\em CVPR}, 2017.

\bibitem{boykov:2001}
Y.~Boykov, O.~Veksler, and R.~Zabih.
\newblock Fast approximate energy minimization via graph cuts.
\newblock {\em PAMI}, 2001.

\bibitem{bregler:2000}
C.~Bregler, A.~Hertzmann, and H.~Biermann.
\newblock Recovering non-rigid 3{D} shape from image streams.
\newblock In {\em CVPR}, 1999.

\bibitem{brox:2010}
T.~Brox and J.~Malik.
\newblock Object segmentation by long term analysis of point trajectories.
\newblock In {\em ECCV}, 2010.

\bibitem{dai:2012}
Y.~Dai, H.~Li, and M.~He.
\newblock A simple prior-free method for non-rigid structure-from-motion
  factorization.
\newblock In {\em CVPR}, 2012.

\bibitem{bue:2007}
A.~{Del Bue}, X.~Llad\'o, and L.~Agapito.
\newblock Segmentation of rigid motion from non-rigid 2d trajectories.
\newblock {\em Pattern Recognition and Image Analysis}, 2007.

\bibitem{delong:2012}
A.~Delong, A.~Osokin, H.~N. Isack, and Y.~Boykov.
\newblock Fast approximate energy minimization with label costs.
\newblock {\em IJCV}, 2012.

\bibitem{elhamifar:2009}
E.~Elhamifar and R.~Vidal.
\newblock Sparse subspace clustering: Algorithm, theory, and applications.
\newblock In {\em CVPR}, 2009.

\bibitem{fischler:1981}
M.~A. Fischler and R.~C. Bolles.
\newblock Random sample consensus: A paradigm for model fitting with
  applications to image analysis and automated cartography.
\newblock {\em ACM Communications}, 1981.

\bibitem{nrsfmFragkiadaki}
K.~Fragkiadaki, M.~Salas, P.~Arbelaez, and J.~Malik.
\newblock Grouping-based low-rank trajectory completion and 3d reconstruction.
\newblock In {\em NIPS}, 2014.

\bibitem{fragkiadaki:2012}
K.~Fragkiadaki, G.~Zhang, and J.~Shi.
\newblock Video segmentation by tracing discontinuities in a trajectory
  embedding.
\newblock In {\em CVPR}, 2012.

\bibitem{hartley:2004}
R.~Hartley and A.~Zisserman.
\newblock {\em Multiple View Geometry in Computer Vision}.
\newblock Cambridge University Press, second edition, 2004.

\bibitem{joo_iccv_2015}
H.~Joo, H.~Liu, L.~Tan, L.~Gui, B.~Nabbe, I.~Matthews, T.~Kanade, S.~Nobuhara,
  and Y.~Sheikh.
\newblock Panoptic studio: A massively multiview system for social motion
  capture.
\newblock In {\em ICCV}, 2015.

\bibitem{joo_cvpr_2014}
H.~Joo, H.~S. Park, and Y.~Sheikh.
\newblock Map visibility estimation for large-scale dynamic 3d reconstruction.
\newblock In {\em CVPR}, 2014.

\bibitem{kaminski:2004}
J.~Y. Kaminski and M.~Teicher.
\newblock A general framework for trajectory triangulation.
\newblock {\em Journal of Mathematical Imaging and Vision}, 2004.

\bibitem{kundu:2014}
A.~Kundu, Y.~Li, F.~Daellert, F.~Li, and J.~M. Rehg.
\newblock Joint semantic segmentation and 3d reconstruction from monocular
  video.
\newblock In {\em ECCV}, 2014.

\bibitem{kundu:2016}
A.~Kundu, V.~Vineet, and V.~Koltun.
\newblock Feature space optimization for semantic video segmentation.
\newblock In {\em CVPR}, 2016.

\bibitem{lin:2017}
G.~Lin, A.~Milan, C.~Shen, and I.~Reid.
\newblock Refinenet: Multi-path refinement networks for high-resolution
  semantic segmentation.
\newblock In {\em CVPR}, 2017.

\bibitem{ozden:2004}
K.~E. Ozden, K.~Cornelis, L.~V. Eychen, and L.~V. Gool.
\newblock Reconstructing 3{D} trajectories of independently moving objects
  using generic constraints.
\newblock {\em CVIU}, 2004.

\bibitem{park:2011}
H.~S. Park and Y.~Sheikh.
\newblock 3d reconstruction of a smooth articulated trajectory from a monocular
  image sequence.
\newblock In {\em ICCV}, 2011.

\bibitem{park:2010}
H.~S. Park, T.~Shiratori, I.~Matthews, and Y.~Sheikh.
\newblock 3{D} reconstruction of a moving point from a series of 2{D}
  projections.
\newblock In {\em ECCV}, 2010.

\bibitem{rao:2010}
S.~Rao, R.~Tron, R.~Vidal, and Y.~Ma.
\newblock Motion segmentation in the presence of outlying, incomplete, or
  corrupted trajectories.
\newblock {\em PAMI}, 2010.

\bibitem{redmon:2017}
J.~Redmon and A.~Farhadi.
\newblock Yolo9000: Better, faster, stronger.
\newblock In {\em CVPR}, 2017.

\bibitem{ricco:2013}
S.~Ricco and C.~Tomasi.
\newblock Video motion for every visible point.
\newblock In {\em ICCV}, 2013.

\bibitem{russell:2014}
C.~Russell, R.~Yu, and L.~Agapito.
\newblock Video pop-up: Monocular 3d reconstruction of dynamic scenes.
\newblock In {\em NIPS}, 2014.

\bibitem{salzmann:2007}
M.~Salzmann, J.~Pilet, S.~Ilic, and P.~Fua.
\newblock Surface deformation models for nonrigid 3{D} shape recovery.
\newblock {\em PAMI}, 2007.

\bibitem{shaji:2010}
A.~Shaji, A.~Varol, L.~Torresani, and P.~Fua.
\newblock Simulataneous point matching and 3{D} deformable surface
  reconstruction.
\newblock In {\em CVPR}, 2010.

\bibitem{sheikh:2009}
Y.~Sheikh, O.~Javed, and T.~Kanade.
\newblock Background subtraction for freely moving cameras.
\newblock In {\em ICCV}, 2009.

\bibitem{sidenbladh:2000}
H.~Sidenbladh, M.~J. Black, and D.~J. Fleet.
\newblock Stochastic tracking of 3d human figures using 2{D} image motion.
\newblock In {\em ECCV}, 2000.

\bibitem{SNAVELY-IJCV08}
N.~Snavely, S.~M. Seitz, and R.~Szeliski.
\newblock Modeling the world from {Internet} photo collections.
\newblock {\em IJCV}, 2008.

\bibitem{taylor:2013}
B.~Taylor, A.~Ayvaci, A.~Ravichandran, and S.~Soatto.
\newblock Semantic video segmentation from occlusion relations within a convex
  optimization framework.
\newblock In {\em CVPR Workshop}, 2013.

\bibitem{taylor:2010}
J.~Taylor, A.~D. Jepson, and K.~N. Kutulakos.
\newblock Non-rigid structure from locally-rigid motion.
\newblock In {\em CVPR}, 2010.

\bibitem{torresani:2002}
L.~Torresani and C.~Bregler.
\newblock Space-time tracking.
\newblock In {\em ECCV}, 2002.

\bibitem{torresani:2001}
L.~Torresani, D.~Yang, G.~Alexander, and C.~Bregler.
\newblock Tracking and modeling non-rigid objects with rank constraints.
\newblock In {\em CVPR}, 2001.

\bibitem{yan:2006}
J.~Yan and M.~Pollefeys.
\newblock Automatic kinematic chain building from feature trajectories of
  articulated objects.
\newblock In {\em CVPR}, 2006.

\end{thebibliography}
}
\end{document}